%% file: main.tex
\RequirePackage[svgnames]{xcolor}

\documentclass[11pt,letterpaper]{mystyle}

\usepackage[all]{hypcap}
\usepackage[svgnames]{xcolor}
\usepackage[comma,authoryear,compress]{natbib}
\usepackage{hyperref}[citecolor=lightblue]

\hypersetup{
    colorlinks = true,
    citecolor = {YaleBlue},
}

\usepackage{amssymb}
\usepackage{mathrsfs}
\usepackage{dsfont}
\usepackage{algorithm}
\usepackage{algorithmicx}
\usepackage{algpseudocode}
\usepackage{microtype}
\usepackage{graphicx}
\expandafter\def\csname ver@subfig.sty\endcsname{}
\usepackage{booktabs} %
\usepackage{float}
\usepackage{bigstrut}

\usepackage{amsmath}

\usepackage{mathtools}
\usepackage{amsthm}

\usepackage{nicefrac}

\usepackage{enumitem}
\usepackage{subcaption}
\usepackage{graphicx,subfig}
\usepackage{cleveref}
\usepackage{bxcoloremoji}

\usepackage{float}

\usepackage[utf8]{inputenc} %
\usepackage[T1]{fontenc}    %
\usepackage{hyperref}       %
\usepackage{url}            %
\usepackage{booktabs}       %
\usepackage{amsfonts}       %
\usepackage{nicefrac}       %
\usepackage{microtype}      %
\usepackage{graphicx}
\usepackage{amssymb}
\usepackage{fdsymbol}
\usepackage{wrapfig}
\usepackage{enumitem}
\usepackage{stackengine}
\usepackage[font=small,labelfont=bf]{caption}
\usepackage{color}
\usepackage{adjustbox}

\usepackage{rotating}
\usepackage{makecell}

\usepackage{multirow}

\input{macro}

\newtcolorbox{AIbox}[2][]{aibox,title=#2,#1}
\definecolor{lightblue}{rgb}{0.22,0.45,0.70}%
\definecolor{Gray}{gray}{0.95}
\definecolor{Cornsilk}{rgb}{1.0, 0.97, 0.86}

\usepackage{pifont}
\usepackage{soul}

\usepackage[all]{hypcap}
\usepackage{forest}

\usepackage{wasysym}
\usepackage[utf8]{inputenc}
\usepackage[T1]{fontenc}

\usepackage{mathpazo}
\usepackage{tcolorbox}

\usepackage{bbding}

\newtcolorbox{simpleElegantQuote}{
    colback=AliceBlue!50!White,   
    colframe=RoyalBlue!75!Black,  
    boxrule=0.5pt,                
    arc=2mm,                      
    boxsep=4pt,                   
    left=10pt, right=10pt,        
    top=8pt, bottom=8pt,          
    fontupper=\itshape,           
}

\title{\paperlogo{} How Far Are AI Scientists from \\Changing the World?}

\runningtitle{\paperlogo{} How Far Are AI Scientists from Changing the World?}

\author{
  Qiujie Xie*$^{1,2}$, Yixuan Weng*$^1$, Minjun Zhu*$^{1,2}$, \\
  Fuchen Shen$^1$, Shulin Huang$^1$, Zhen Lin$^1$, Jiahui Zhou$^3$, Zilan Mao$^1$, \\
  Zijie Yang$^{1,4}$, Linyi Yang$^5$, Jian Wu$^1$, Yue Zhang\dag
}

\affil[1]{School of Engineering, Westlake University}
\affil[2]{Zhejiang University}
\affil[3]{School of Software, Dalian University of Technology}
\affil[4]{School of Life Sciences, Westlake University}
\affil[5]{University College London}

\correspondingauthor{Yue Zhang: \href{mailto:zhangyue@westlake.edu.cn}{zhangyue@westlake.edu.cn}}

\begin{document}

\begin{abstract}
The emergence of large language models (LLMs) is propelling automated scientific discovery to the next level, with LLM-based Artificial Intelligence (AI) Scientist systems now taking the lead in scientific research. Several influential works have already appeared in the field of AI Scientist systems, with AI-generated research papers having been accepted at the ICLR 2025 workshop, suggesting that a human-level AI Scientist capable of uncovering phenomena previously unknown to humans, may soon become a reality. In this survey, we focus on the central question: How far are AI scientists from changing the world and reshaping the scientific research paradigm? To answer this question, we provide a prospect-driven review that comprehensively analyzes the current achievements of AI Scientist systems, identifying key bottlenecks and the critical components required for the emergence of a scientific agent capable of producing ground-breaking discoveries that solve grand challenges. We hope this survey will contribute to a clearer understanding of limitations of current AI Scientist systems, showing where we are, what is missing, and what the ultimate goals for scientific AI should be.

\vspace{2mm}


\vspace{5mm}

\coloremojicode{1F4C5} \textbf{Date}: August 1st, 2025

\github{} \textbf{Code Repository}: \href{https://github.com/ResearAI/Awesome-AI-Scientist}{https://github.com/ResearAI/Awesome-AI-Scientist}

\coloremojicode{1F4E7} \textbf{Contact}: \href{mailto:zhangyue@westlake.edu.cn}{zhangyue@westlake.edu.cn}

\end{abstract}

\maketitle
\vspace{3mm}
\section{Introduction}
\label{sec:intro}

Scientific discovery is the fundamental driving force behind the advancement of human civilization. From Newton's laws of motion and gravitation~\citep{newton1833philosophiae, newton1850newton}, through Einstein's theory of relativity~\citep{einstein1922general}, to artificial intelligence~\citep{lecun2015deep,ertel2024introduction}, every major scientific breakthrough has expanded the boundaries of human understanding and propelled societal progress. Traditionally, humans, as the primary agents of scientific research, have followed a systematic path of exploration. This process typically begins with observation and learning to establish a foundational basis for scientific knowledge \textbf{(knowledge acquisition)}, followed by the formulation of scientific hypotheses to address unresolved questions \textbf{(idea generation)}. The subsequent step involves rigorous testing and falsification of these hypotheses through intensive experiments and theoretical analysis \textbf{(verification and falsification)}. Finally, research is continuously refined based on experimental or theoretical results, driving the ongoing \textbf{evolution} of scientific knowledge~\citep{langley1987scientific}.

\begin{figure*}[tbp]
    \centering
    \includegraphics[width=0.85\linewidth]{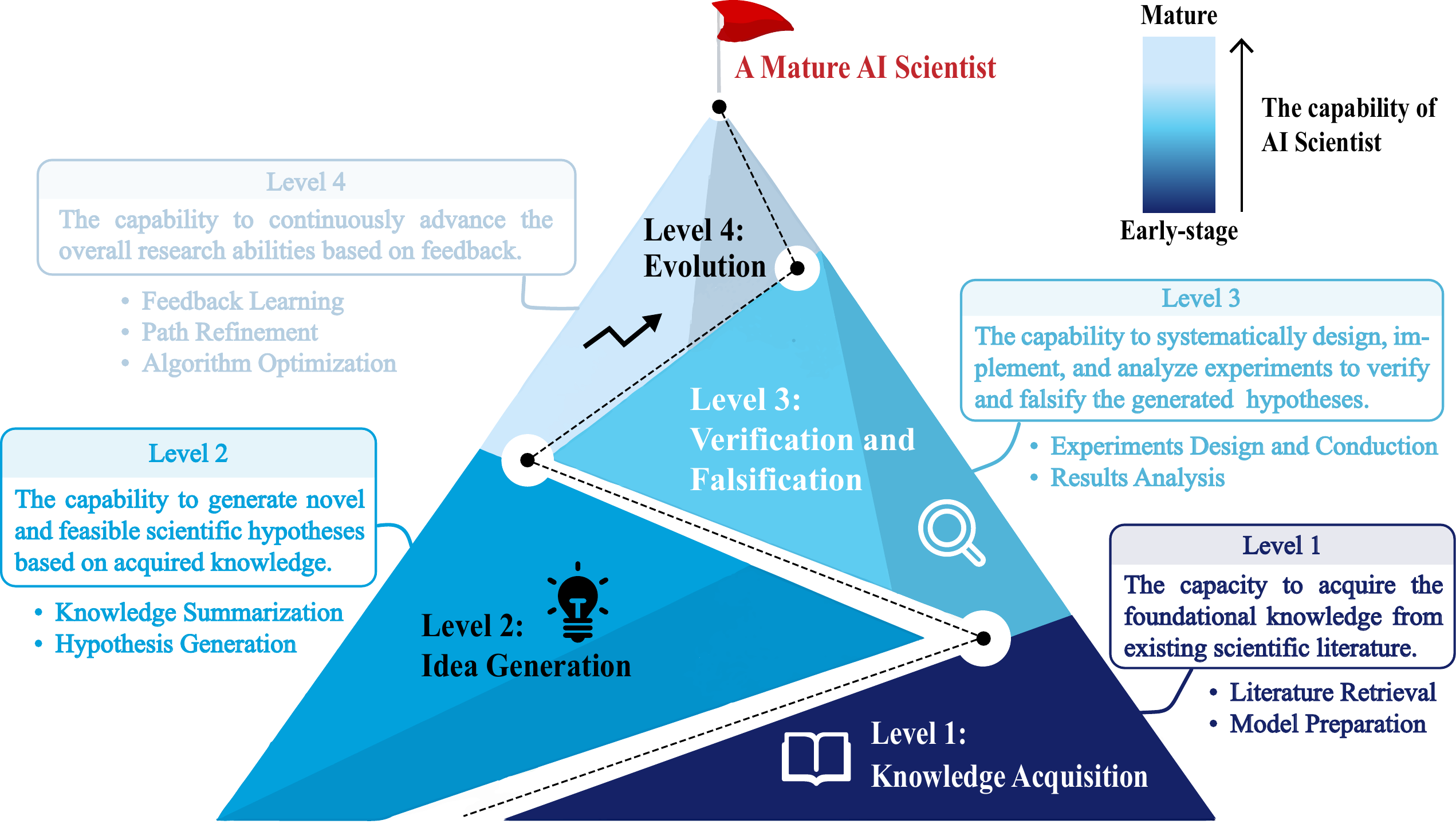}
    \caption{The capability level of an AI Scientist, illustrating the progression from foundational knowledge acquisition (Level 1), through idea generation (Level 2), rigorous hypothesis verification and falsification (Level 3), to continuous evolution (Level 4). We outline the core functions for each capability level.}
    \label{fig:capability-level}
\end{figure*}
\input{figures/main_figure.tex}

However, the research process is inherently constrained by human limitations such as limited time and cognitive capacity, leading to slow literature reviews, relatively narrow knowledge domains, biased hypothesis generation, and inefficient experiment execution~\citep{ioannidis2005most,baker20161}. To address these issues, there has been a growing pursuit of the automation of scientific discovery~\citep{yang2023ai,yang2025airalogyaiempowereduniversaldata}. Early efforts have focused on leveraging \textbf{deep learning techniques}~\citep{lecun2015deep, vaswani2017attention} to 
support knowledge acquisition. For example, pre-trained language models have been adapted with domain-specific knowledge to better represent scientific information~\citep{gupta_matscibert_2022}. 
However, due to constraints in model capacity and the availability of high-quality scientific data, these methods have primarily served as scientific tools, providing partial acceleration to the human research process. 
In recent years, \textbf{LLMs trained on large-scale, high-quality corpora} have demonstrated remarkable abilities in both comprehension and generation across long-text scenarios, making it possible to generate novel and feasible scientific hypotheses based on the accumulated knowledge~\citep{si2024can}, and to rigorously evaluate these hypotheses using automated experimental tools~\citep{wierenga2023pylabrobot, starace2025paperbench}. This progress is propelling automated scientific discovery to the next level, where LLM-based AI Scientist systems~\citep{lu2024ai, weng2025cycleresearcher} are now taking the lead in exploration. 
Once endowed with capabilities beyond human limitations~(e.g., the ability to overcome restricted memory capacity), an AI Scientist will have the potential to make ground-breaking discoveries that solve grand challenges across medicine, energy, and the environment.
We are now poised to explore a central question in the field: \textbf{How far are AI scientists from changing the world and reshaping the scientific research paradigm?}

To answer this question, we provide a prospect-driven review of existing achievements. Figure~\ref{fig:capability-level} lays out a hierarchy of capacities necessary for a mature AI Scientist, which we reckon as a reasonable road map for current and future research in this direction. Specifically, \textbf{our survey introduces a capability-level framework that systematically defines the stages of AI scientist development, including: } (1) knowledge acquisition, which is \textbf{the foundational capability} of an AI Scientist, encompassing the autonomous capability to retrieve, review, and comprehend scientific knowledge from existing research. (2) Idea generation, refers to the capability to generate innovative and feasible hypotheses at scale. This capability serves as the key feature \textbf{distinguishing AI scientist systems from automated scientific tools}. (3) Verification and falsification, the capability to systematically design, implement, and analyze experiments to test and potentially disprove the AI-generated scientific hypotheses, transforming the AI Scientist from an idea generator into \textbf{an autonomous scientific intelligence}. (4) Evolution, the capability to continuously advance overall research abilities based on internal and external feedback, which is essential for an elementary AI Scientist to \textbf{evolve into a mature scientific agent}. We critically analyze current research through the lens of this capability framework, identifying key bottlenecks and missing components necessary for the emergence of ground-breaking discoveries produced by autonomous scientific intelligence.

For the above capabilities, there have been endeavors, successes, and limitations~(Figure~\ref{fig:introduction}). 
We first systematically review existing achievements in knowledge acquisition~(Sections~\ref{sec:knowledge_acquisition}), idea generation~(Sections~\ref{sec:idea_generation}), and verification and falsification~(Sections~\ref{sec:verification_and_falsification}) capabilities of current AI Scientist systems, mapping representative methods to each capability level. We then outline the development of AI reviewer systems and present empirical evidence demonstrating that current AI Scientist systems lack the ability to independently conduct high-quality scientific discovery~(Section~\ref{sec:review}). To overcome these limitations, it is essential for AI Scientist systems to possess the capability for evolution (Section~\ref{sec:evolution}), thus moving toward the goal of reshaping the scientific research paradigm. Furthermore, we comprehensively examine the challenges faced in current AI Scientist research~(Section~\ref{sec:discussion}) and discuss future directions and open questions~(Section~\ref{sec:future_and_conclusion}), offering a reasonable road map toward truly autonomous scientific intelligence.

\section{Knowledge Acquisition}
\label{sec:knowledge_acquisition}
\input{sub_tex/knowledge_acquisition}

\section{Idea Generation}
\label{sec:idea_generation}
\input{sub_tex/idea_generation}

\section{Verification and Falsification}
\label{sec:verification_and_falsification}
\input{sub_tex/verification_and_falsification}

\section{Review System}
\label{sec:review}
\input{sub_tex/review}

\section{Evolution}
\label{sec:evolution}
\input{sub_tex/evolution.tex}

\section{Discussion}
\label{sec:discussion}
\input{sub_tex/discussion}

\section{Future Directions}
\label{sec:future_and_conclusion}
\input{sub_tex/future_and_conclusion}

\section{Conclusion}
This paper presented a systematic survey of existing research on AI Scientist systems, providing a comprehensive overview of advances in the field. Specifically, we proposed a capability-level framework that divides the capabilities of an AI Scientist into four progressive levels: knowledge acquisition, idea generation, verification and falsification, and evolution. Based on this framework, we critically analyzed current research and highlighted gaps AI Scientist systems must address before they can make ground-breaking discoveries that solve grand challenges across medicine, energy, and the environment, thereby changing the world and reshaping the scientific research paradigm. Finally, we concluded the survey by outlining future directions essential for closing the gap and advancing the field toward truly autonomous scientific intelligence.

\clearpage
\bibliography{main}

\end{document}

%% file: macro.tex
\newcommand{\x}{\mathbf{x}}

\definecolor{blanchedalmond}{rgb}{1.0, 0.92, 0.8}
\definecolor{carmine}{rgb}{0.59, 0.0, 0.09}
\definecolor{lightblue}{rgb}{0.22,0.45,0.70}%

\renewcommand{\mathbf}{\boldsymbol}

\makeatletter
\def\Ddots{\mathinner{\mkern1mu\raise\p@
\vbox{\kern7\p@\hbox{.}}\mkern2mu
\raise4\p@\hbox{.}\mkern2mu\raise7\p@\hbox{.}\mkern1mu}}
\makeatother

\definecolor{amaranth}{rgb}{0.9, 0.17, 0.31}
\definecolor{antiquebrass}{rgb}{0.8, 0.58, 0.46}
\definecolor{antiquefuchsia}{rgb}{0.57, 0.36, 0.51}
\definecolor{chromeyellow}{rgb}{0.31, 0.47, 0.26}

\newcommand{\github}{\raisebox{-1.5pt}{\includegraphics[height=1.05em]{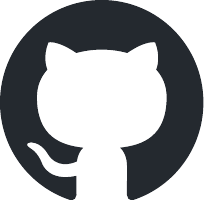}}}

\newcommand{\paperlogo}{\raisebox{-1.5pt}{\includegraphics[height=2.05em]{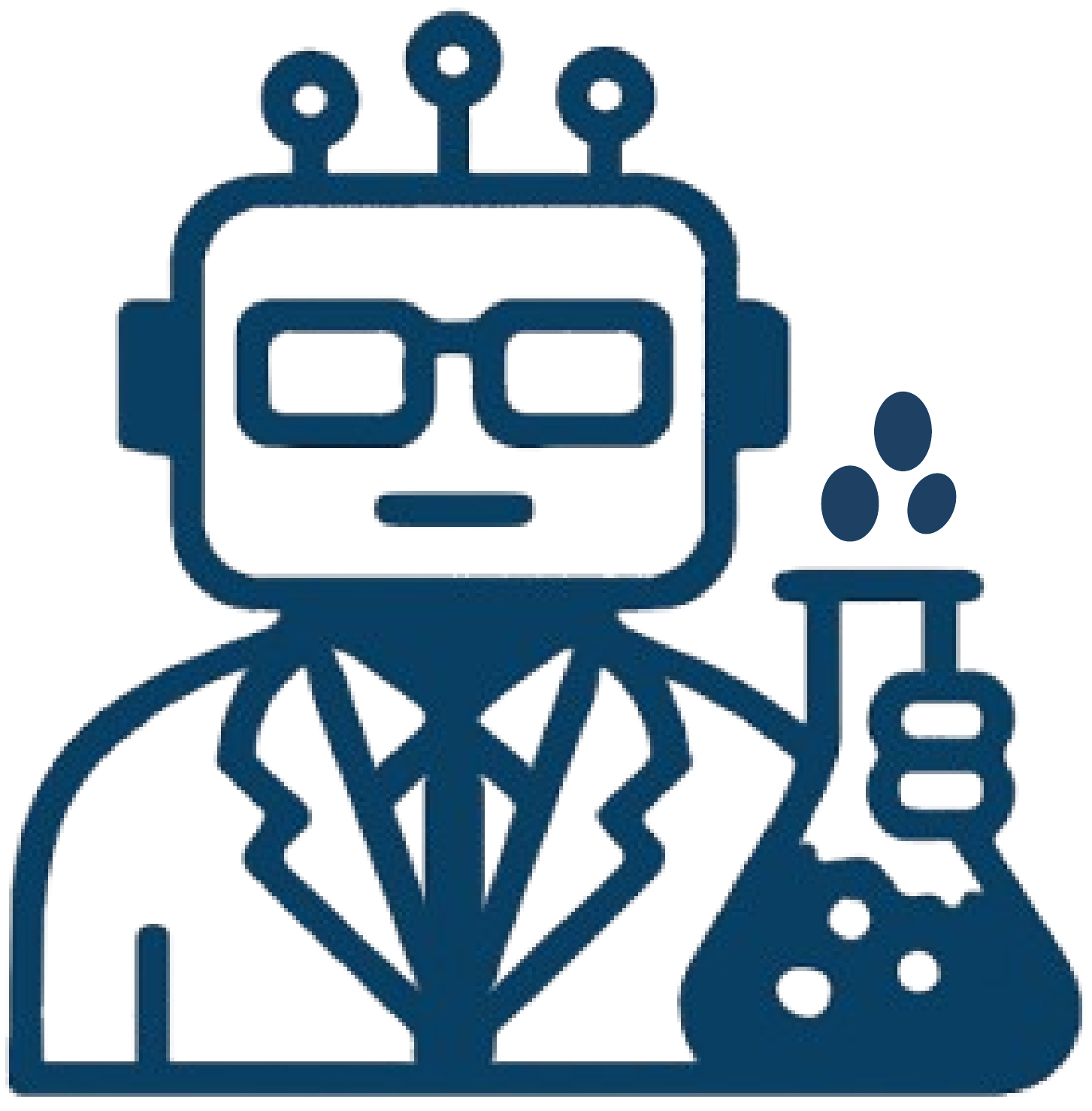}}}

%% file: figures/main_figure.tex
\begin{figure*}[htbp]
    \centering
    \includegraphics[width=1\linewidth]{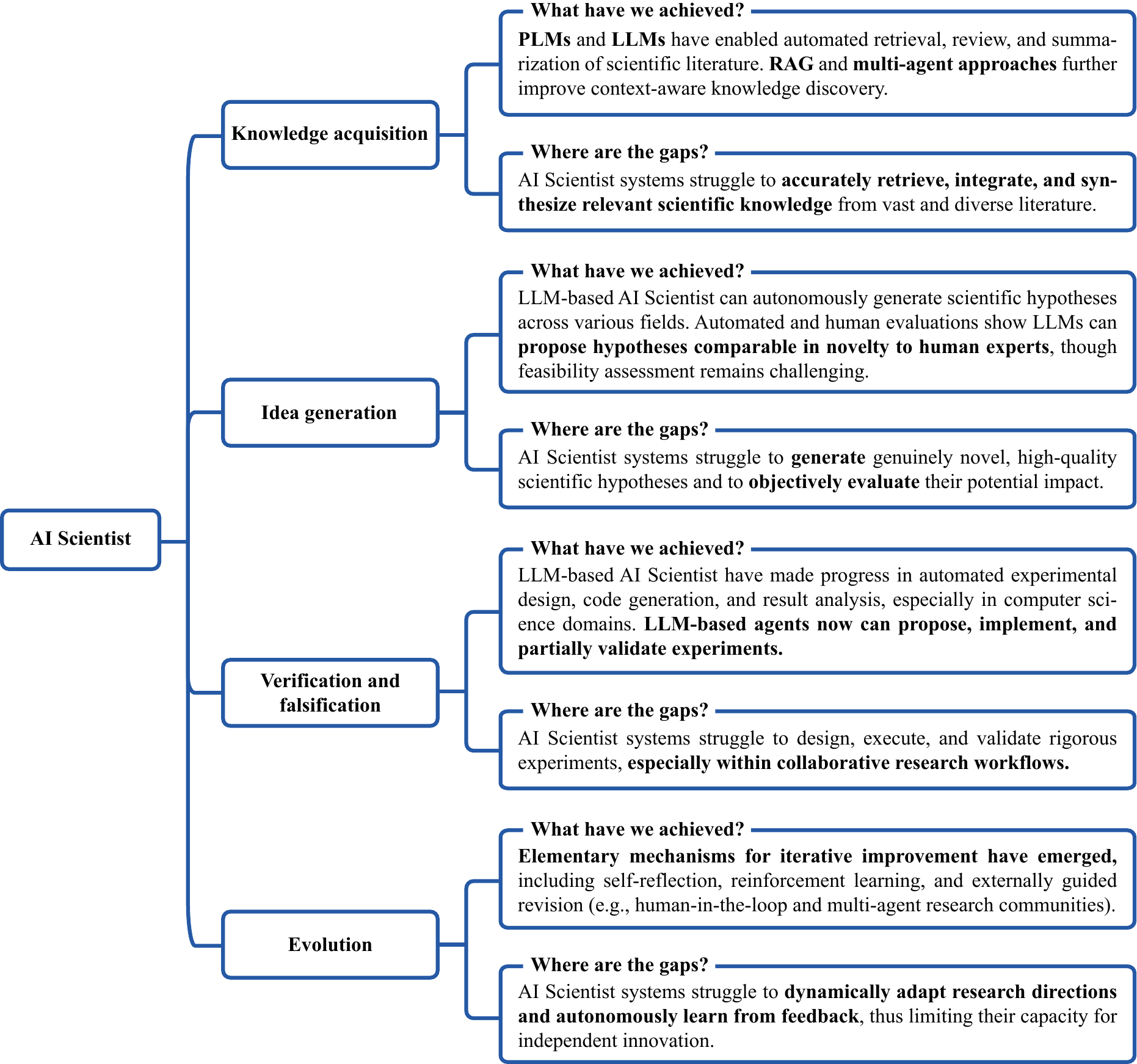}
    \caption{The current capability landscape of AI Scientist systems across four progressive levels. We summarize the current achievements for each level and highlight critical gaps before AI Scientist systems can autonomously make ground-breaking scientific discoveries.}
    \label{fig:introduction}
\end{figure*}

%% file: sub_tex/knowledge_acquisition.tex
\begin{simpleElegantQuote}
\textbf{Definition:} Knowledge acquisition is the foundational capability of an AI Scientist, defined as its autonomous capability to retrieve, review, and comprehend domain-specific knowledge from existing scientific literature, thereby systematically establishing a scientific knowledge base for subsequent research.
\end{simpleElegantQuote}
Knowledge acquisition represents the foundational capability of an AI Scientist, encompassing \textbf{the autonomous capability to retrieve, review, and comprehend domain-specific knowledge from existing scientific research.}
This critical capability establishes the knowledge base necessary for subsequent research activities, functioning similarly to human researchers' literature review processes. We systematically examine how modern AI Scientist systems, particularly LLM-powered methods, address the dual challenges of (1) literature search/curation (filtering relevant research from massive corpora) and (2) knowledge retrieval/summarization (extracting structured insights). Interestingly, even before the advent of LLMs, the foundational concept of AI Scientist systems had already garnered significant attention, with numerous smaller-scale language models demonstrating notable success in the knowledge acquisition phase. Consequently, in this section, we categorize research work of AI Scientist systems in knowledge acquisition into the \textbf{pre-LLM era} (roughly before 2020) and the \textbf{LLM era} (from 2020 onwards, marked by the emergence of LLMs such as GPT-3~\citep{brown2020language}).

\subsection{Pre-LLM Era Methods}
During the pre-LLM era, researchers make significant progress in automated knowledge acquisition by fine-tuning small pre-trained language models (PLMs) with domain-specific knowledge to learn substantial scientific representations~\citep{Beltagy2019SciBERT, jin2019probing, Li2023AllDO}. These models demonstrate remarkable capabilities in structured text mining and knowledge extraction from scientific literature, paving the way for more advanced methods based on LLMs.

\textbf{Literature.}
Prior to the rise of LLMs, early automated knowledge acquisition systems predominantly relied on smaller PLMs that are trained on substantial scientific corpora. SciBERT~\citep{Beltagy2019SciBERT} is the first PLM that is pretrained on 1.14 million papers and achieves strong performance on scientific NER and citation classification tasks, accelerating human researchers' research efficiency, especially in the computer science domain. Similarly, SCHOLARBERT~\citep{hong2023diminishing}, pretrained on 221B tokens of scholarly text, largely enhances contextual understanding in scientific literature. 
Domain-specific models, such as CinicalBERT~\citep{clinicalbert}, PubMedBERT~\citep{pubmedbert}, BioELMo~\citep{jin2019probing}, BioBERT~\citep{lee2020biobert}, BioMegatron~\citep{shin-etal-2020-biomegatron}, BioM-Transformers~\citep{alrowili-shanker-2021-biom}, and MatSciBERT~\citep{gupta_matscibert_2022} are pretrained on high-quality clinical, biology, and chemistry corpora. These models focus on structured text mining, extracting key phrases, entities, and relationships from scientific abstracts or full papers using techniques like rule-based parsing and graph-based citation analysis~\citep{cachola-etal-2020-tldr}. \cite{whatyouread} leverage the author network graph methods to disclose the deep connection behind users' interaction with the papers. Similarly, \cite{kang2023comlittee} construct a committee of authors with user-selected paper seeds while accepting signals from multiple sources to dynamically discover related literature. While the above pre-LLM era studies achieve notable success in automated knowledge acquisition, their scalability and contextual understanding are fundamentally limited, setting the stage for more sophisticated approaches to knowledge acquisition.

\textbf{Experiments. }
In pre-LLM era, researchers also explore the extraction of knowledge from experimental results (e.g., the construction of leaderboards). 
Following the efforts of utilizing data sources like NLP-progress~\footnote{https://nlpprogress.com/} and Papers with Code~\footnote{https://paperswithcode.com/},  SciGen~\citep{moosavi2021scigen} and Numerical-NLG~\citep{suadaa-etal-2021-towards} are two benchmarks that assess the capability of language models in scientific table description generation. 
However, these works are found to lack stringent quality assurance measures. For instance, there is no standardization of scientific entities across various leaderboards, and the coverage of relevant publications is incomplete. Instead, SCICM~\citep{Li2023AllDO} opts for arXiv, which is an open-access archive for nearly 2.4 million scholarly articles in different domains, providing a large amount of relevant publications. 
Similar to SCICM, TDMS-IE~\citep{hou-etal-2019-identification}, and AxCell~\citep{kardas-etal-2020-axcell} extract ``Task-Dataset-Model'' triples, along with the experiment result entities to build leaderboards automatically. This line of work is extended by subsequent methods such as TELIN~\citep{yang-etal-2022-telin} and ORKG‑Leaderboards~\citep{KABENAMUALU2023ORKGLeaderboardsAS}, marking a clear trend toward improving research efficiency.

\subsection{LLM Era Methods}
\label{LLM_era}
The transformative impact of LLMs on the knowledge acquisition capabilities of AI Scientist systems marks the beginning of the ``LLM era''. These advancements span literature search toolkits, idea- and experiment-oriented knowledge extraction techniques, and specialized evaluation benchmarks. At the same time, they underscore persistent challenges in retrieval precision and multi-document summarization.

\subsubsection{Literature Search and Curation}

Researchers have developed specialized toolkits and systems to efficiently filter, select, and recommend relevant scientific literature from diverse sources (e.g., PubMed~\footnote{https://pubmed.ncbi.nlm.nih.gov/}, arXiv~\footnote{https://arxiv.org/}). Among these works, systems such as PaperWeaver~\citep{lee2024paperweaver} stand out for their versatility, as they are capable of supporting multiple substages throughout the literature search pipeline. Building on this progress, DORA AI Scientist~\citep{naumov2025dora} further advances the field by incorporating Retrieval-Augmented Generation (RAG)~\citep{gao2023retrieval} into its core search process, thereby enhancing contextual relevance and automation. Despite these advancements, not all systems are fully autonomous. For example, semi-automatic frameworks like CodeScientist~\citep{jansen2025codescientist} still rely on user-supplied paper lists for ideation, highlighting the ongoing challenges in achieving truly automated and high-precision literature discovery.

Building on this spectrum of approaches, researchers have developed varied implementations for searching and filtering scientific literature. 
\citet{schmidgall2025agent} utilize the arXiv API to retrieve papers and perform summarization, full-text extraction, and curation tasks. Meanwhile,
\citet{airesearcher} apply citation-based filtering to top-cited domain papers from arXiv. Beyond single-agent retrieval, multi-agent systems have been explored as well. 
\citet{gottweis2025towards} integrate web search tools, while
\citet{lu2024ai} utilize Semantic Scholar’s API to identify relevant papers for hypotheses generation. In the RAG domain, \citet{lala2023paperqa} and \citet{skarlinski2024language} introduce PaperQA and PaperQA2 for scientific question answering tasks. Furthermore, the pursuit of greater autonomy is exemplified by PaSa~\citep{he2024pasa}, which deploys a dual-agent architecture with session-level reinforcement learning to autonomously search, analyze papers, and explore citation networks for complex queries. Despite these advancements, \citet{beel2025evaluating} note that existing systems still heavily rely on keyword-based matching, often retrieving classic but potentially outdated papers, highlighting the need for more context-aware retrieval strategies.

\subsubsection{Knowledge Retrieval and Summarization}
Based on the well-processed literature, knowledge retrieval and summarization methods aim to derive knowledge from both structured and unstructured data. This line of work primarily consists of two parts: (1) idea-oriented approaches, which extract key concepts and insights from research papers, and (2) experiment-oriented approaches, which retrieve experimental results and summarize them into comprehensive reports. LLMs play a pivotal role in this stage by processing and organizing valuable knowledge from the literature.

\textbf{Idea-oriented approaches } utilize LLMs to extract key concepts and insights from scientific documents. These methods can be classified into several categories: 
(1)~\textbf{Direct prompting} serves as a foundational approach, in which the LLM-based task execution is directly guided by explicit prompt instructions. However, due to constraints such as context length and retrieval accuracy, these approaches are rarely adopted and typically serve as the baseline methods~\citep{dagdelen2024structured,gupta2024data,yao2025extracting}. 
(2)~\textbf{Literature structure-based methods} leverage the inherent organization of scientific documents or predefined plans based on paper structures (e.g., abstract, introduction, and conclusion~(AIC))~\citep{sharma2019bigpatent, cachola2020tldr,takeshita2022x}. 
(3)~\textbf{RAG-based methods} enhance the summarization ability of LLM by first retrieving relevant external information and then using it as context for generation, aiming for more factual and timely outputs~\citep{ali2024automated, agarwal2024llms}. For instance, LitLLM~\citep{agarwal2024litllm} utilizes RAG for both literature searching and knowledge retrieval. (4)~\textbf{Multi-agent-based methods} involve the collaboration of multi-agent systems to generate literature reviews. For example, \citet{schmidgall2025agentrxiv} integrate multiple LLMs to accelerate the process of knowledge retrieval and summarization.

\textbf{Experiment-oriented methods }
involve extracting experimental results from scientific documents and summarizing them into comprehensive reports. LLMs play a pivotal role in this process by analyzing structured data, interpreting experimental designs, and generating comprehensive summaries. However, this task is complicated by the need for precise extraction of Task-Model-Dataset triples and the dynamic nature of research benchmarks.
For example, the LEGO-bench~\citep{singh2024legobench} proposes a benchmark for evaluating systems that generate scientific leaderboards, which attempts to stay informed about the latest state-of-the-art research. Through semi-automated extraction of experimental data from multiple datasets, \citet{park2025can} demonstrate how comparative approaches can yield new scientific discoveries. Concurrently, \citet{wu2025lag} propose Leader Auto Generation~(LAG), which implements a comprehensive four-stage process for dynamic leaderboard generation, encompassing table retrieval, data integration, leaderboard generation, and quality-based selection of the most effective leaderboards.

\subsection{Evaluation}
Evaluation benchmarks play a crucial role in assessing the capabilities of AI scientist systems in knowledge acquisition. These benchmarks focus on different facets, ranging from information retrieval accuracy and summarization quality to knowledge structure generation. 

\textbf{Classical scientific summarization.} The early evaluation of knowledge acquisition capability derives from works of literature summarization~\citep{cachola2020tldr,takeshita2022x}. Including more than 5400 summaries for scientific papers, SCITLDR~\citep{cachola2020tldr} requires models to output extremely compressed summarizations of scientific papers. Following the same format as SCITLDR, X-SCITLDR~\citep{cachola2020tldr} is a multilingual dataset that tests the effectiveness of different cross-lingual summarization strategies. These benchmarks only utilize parts of the original scientific papers~(e.g., abstract) as testing items, thus aiming at models with relatively shorter context windows.

\textbf{Literature retrieval}. Beyond summarization, another critical dimension of knowledge acquisition evaluation lies in a model's ability to retrieve relevant scientific literature. CiteME~\citep{press2024citeme} is a recent benchmark focusing on evaluating LLM-based scientific agents by identifying whether they can cite the correct papers based on given context. LitSearch~\citep{ajith2024litsearch} is another benchmark that assesses the ability of scientific literature retrieval systems, containing 597 manually curated queries and 64k+ papers to test the ability of complex query comprehension and relevant document location.

\textbf{Automated literature review and tools. }
Recent benchmarks have started to evaluate the ability of LLMs to perform automated literature review and assist in scientific knowledge organization. For example, \citet{yun2023appraising} find that researchers tend to prefer transparent AI tools over black-box systems for literature review tasks. Building on this, \citet{hsu-etal-2024-chime} introduce CHIME, a semi-automatically generated benchmark that assesses the ability of LLMs to generate and link topic categories, though these models still struggle with assigning concrete studies to appropriate topics. Similarly, \citet{agarwal2024llms} propose new evaluation protocols and test sets derived from arXiv papers to rigorously test whether LLMs can write literature reviews, with a focus on zero-shot integrity and avoiding test set contamination as models evolve.

%% file: sub_tex/idea_generation.tex
\begin{simpleElegantQuote}
\textbf{Definition:} 
The idea generation capability of an AI scientist lies in its capability to autonomously formulate innovative and feasible scientific hypotheses based on acquired knowledge and existing methods. \textbf{This capability distinguishes AI scientist systems from automated scientific tools.}
\end{simpleElegantQuote}

In the traditional human-centric research paradigm, researchers are responsible for formulating ideas to be tested, while AI handles labor-intensive tasks such as idea implementation and refinement. In this paradigm, AI is regarded as an automated tool, rather than a principal actor in scientific research. The emergence of LLMs marks a shift, as AI can now autonomously retrieve, review, and comprehend domain-specific knowledge from vast bodies of scientific literature (Section~\ref{sec:knowledge_acquisition}). 
As a result, an increasing number of efforts are being devoted to developing AI scientist systems capable of autonomously generating innovative and feasible hypotheses at scale~\citep{hu2024nova, lu2024ai, si2024can, weng2025cycleresearcher}. 
In this new paradigm, AI scientist systems, distinguished from automated scientific tools, \textbf{become the core drivers of scientific research, responsible for proposing novel hypotheses,} while humans or automated experimental tools act as the executors of these hypotheses.

\subsection{Methods}
A classic formalization of idea generation dates back to \citet{swanson1986undiscovered}, who proposed the ``ABC'' model: two concepts, A and C, are hypothesized to be linked if they both co-occur with some intermediate concept B in the literature. Inspired by this work, subsequent studies have adopted \textbf{graph models}~\citep{wang2019paperrobot, sybrandt2020agatha, youn2022knowledge} or \textbf{word vectors}~\citep{tshitoyan2019unsupervised} to generate scientific hypotheses by identifying links between concept pairs. However, transforming complex scientific ideas into concept links in binary form not only constrains the input type but also significantly limits the expressivity of the generated hypotheses ~\citep{moreau2021literature, wang-etal-2024-scimon}.

In recent years, the rapid advancement of LLMs has opened up new possibilities for 
researchers across various fields to explore how to enable LLMs to generate innovative scientific hypotheses in the form of natural language, through specialized \textbf{prompting} strategies~\citep{li2024chain,hu2024nova,lu2024ai,yamada2025ai,jansen2025codescientist}, \textbf{post-training} methods~\citep{qi2023large, wang-etal-2024-scimon, weng2025cycleresearcher}, or \textbf{multi-agent collaboration}~\citep{liu2024aigs, yang2024large, yang2025moosechem, gottweis2025towards}. For instance, drawing inspiration from how humans conduct research, \citet{li2024chain} propose the Chain-of-Ideas (CoI) agent. This LLM-based agent organizes relevant literature into a chain structure, effectively reflecting the progressive development of a research domain and enabling the generation of high-quality hypotheses grounded in existing literature. Building on the importance of retrieval, \citet{si2024can} employ an LLM agent that incorporates retrieval augmentation and leverages recent advances in inference-time scaling. They prompt the LLM to generate 4,000 seed ideas for each research topic and use a reranker to select the highest-quality hypotheses. Another notable line of work focuses on post-training optimization. For instance, \citet{wang-etal-2024-scimon} introduce SCIMON, a framework designed to generate natural language hypotheses based on background contexts dynamically retrieved from scientific literature. SCIMON further fine-tunes T5 models~\citep{Raffel-T5-2020} using both an in-context contrastive objective and a language modeling objective, and then explicitly optimizing the generated hypotheses for novelty through an iterative process. Meanwhile, \textsc{AI Scientist}~\citep{lu2024ai} employs multiple rounds of chain-of-thought~\citep{wei2022chain} and self-reflection~\citep{shinn2023reflexion} to refine and develop each hypothesis. Beyond individual agents, collaborative multi-agent systems have also demonstrated promise. For example, \citet{yang2025moosechem} develop MOOSE-CHEM, an LLM-based multi-agent framework that operates in three stages: (1) Searching chemistry literature for inspiration papers, (2) using these inspirations to propose hypotheses related to the target research question, and (3) identifying and ranking high-quality hypotheses.

\subsection{Evaluation}

\begin{table}[ht]
\centering
\small
\caption{Commonly-used evaluation criteria for AI-generated scientific hypotheses.}
\label{tab:hypothesis_evaluation}
\renewcommand{\arraystretch}{1}
\begin{tabularx}{0.75\textwidth}{@{}p{0.15\textwidth}p{0.575\textwidth}@{}}
\toprule
\textbf{Criterion} & \textbf{Description} \\
\midrule
\multirow{2}{*}{\textbf{Novelty}} & Whether the hypothesis is creative, distinct from existing work on the topic, and offers fresh insights. \\
\rowcolor[rgb]{ .949,  .949,  .949}
\multirow{2}{*}{\textbf{Feasibility}} & How practical and achievable the hypothesis is as the basis for a research project. \\
\textbf{Clarity} & Whether the hypothesis is clearly stated and easy to understand. \\
\rowcolor[rgb]{ .949,  .949,  .949}
\multirow{2}{*}{\textbf{Excitement}} & The potential excitement and impact the hypothesis could generate if pursued as a full research endeavor. \\
\bottomrule
\end{tabularx}
\end{table}

Evaluation of the generated hypotheses is typically conducted based on multiple criteria that reflect their quality, including but not limited to: novelty, feasibility, clarity, and excitement~\citep{si2024can, chai2024exploring, wang-etal-2024-scimon, li2024chain}. The detail of each criterion is presented in Table~\ref{tab:hypothesis_evaluation}. 
Based on these evaluation criteria, researchers generally adopt two approaches to estimate the quality of AI-generated hypotheses: automated evaluation and human evaluation. Among them, \textbf{automated methods}~\citep{li2024chain, yang2025moosechem, chai2024exploring, yang2024large,chai2024exploring} usually involve employing a powerful LLM as an auto-evaluator to assess hypothesis quality. For instance, \citet{chai2024exploring} employ the Claude 3.5 model~\citep{anthropic2024claude35sonnet} to compare each hypothesis against the background context and research topic, ensuring it demonstrates sufficient novelty, scientific soundness, and clarity. \citet{li2024chain} propose Idea Arena, a pairwise evaluation system in which an LLM judge ranks hypotheses in pairwise comparisons and computes ELO scores for each hypothesis generation model. However, \textbf{the performance of LLMs as evaluators is influenced by various factors}, such as training data diversity~\citep{shidetecting}, inherent model biases~\citep{zheng2023judging}, and evaluation uncertainty~\citep{xie2025an}. 

In contrast, \textbf{human evaluations}~\citep{qi2023large, si2024can, hu2024nova, wang-etal-2024-scimon} are considered the gold standard for assessing hypotheses, as quantitative metrics often fall short. For example, \citet{hu2024nova} recruit a panel of 10 experts, all holding a PhD degree or professorship to evaluate hypotheses based on novelty and overall quality. Furthermore, \citet{si2024can} conduct a large-scale human study that recruits 79 expert researchers to perform blind reviews of 49 ideas from each of three conditions: expert-written ideas, AI-generated ideas, and AI-generated ideas re-ranked by a human expert. They find that LLM-generated ideas are rated as more novel ($p$ < 0.05) than those written by human experts, although they are considered slightly less feasible. Additionally, some studies also utilize \textbf{reference-based text generation metrics} to evaluate hypothesis quality. For example, \citet{qi2023large} employ BLEU~\citep{papineni2002bleu} and ROUGE~\citep{lin2004rouge} scores to measure word overlap between generated outputs and ground truth references. \citet{yang2025moosechem} adopt the Matched Score, which calculates the similarity between a generated hypothesis and the original hypothesis using a 6-point Likert scale.

It is worth noting that most existing approaches to evaluating the feasibility of an AI-generated hypothesis still heavily rely on human intuition and subjective estimation based on textual descriptions. While this method can efficiently filter out obviously infeasible hypotheses, it lacks accuracy and objectivity~\citep{krishna2023longeval, karpinska2021perils}. Therefore, \textbf{assessing the feasibility of a hypothesis also requires practical methods} (e.g., transforming the hypothesis into executable code) to enable empirical validation and falsification. In this survey, we categorize the ability to design experiments to validate new scientific hypotheses as a more advanced form of verification and falsification, which we discuss in detail in Section~\ref{sec:verification_and_falsification}.

%% file: sub_tex/verification_and_falsification.tex
\begin{simpleElegantQuote}
\textbf{Definition}: This capability allows an AI Scientist to systematically design, implement, and analyze experiments to \textbf{verify} and \textbf{falsify} the AI-generated scientific hypotheses. Crucially, the verification and falsification capability is what completes the research cycle, transforming the AI Scientist from an idea generator into an autonomous scientific intelligence.
\end{simpleElegantQuote}

While AI systems show growing proficiency in generating scientific hypotheses, concerns persist about their reliability and true scientific merit. As noted in Section~\ref{sec:idea_generation}, current assessments of these AI-generated hypotheses, especially regarding their \textbf{feasibility}, rely heavily on subjective human evaluations or superficial text-similarity metrics. 
These approaches fundamentally lack the \textbf{empirical rigor} that only practical experimental validation can provide, which underscores the indispensable role of \textbf{verification and falsification capabilities}. The capability to actively test and validate hypotheses (verification) and rigorously challenge them (falsification) in simulated or real environments is the crucial component that defines \textbf{a complete AI Scientist capable of closed-loop scientific discovery.}

\begin{wrapfigure}{r}{0.48\textwidth} 
\vspace{-10pt} 
\centering
\resizebox{0.47\textwidth}{!}{%
\begin{tabular}{ccc}
\toprule
\textbf{Type} & \textbf{Total citations} & \textbf{Avg. citations} \\
\midrule
Without Implementation & 216 & 10.3 \\
\rowcolor[rgb]{ .949,  .949,  .949}
With Implementation & \textbf{325}  & \textbf{25.0} \\
\bottomrule
\end{tabular}%
} 
\vspace{5pt} 
\includegraphics[width=0.95\linewidth]{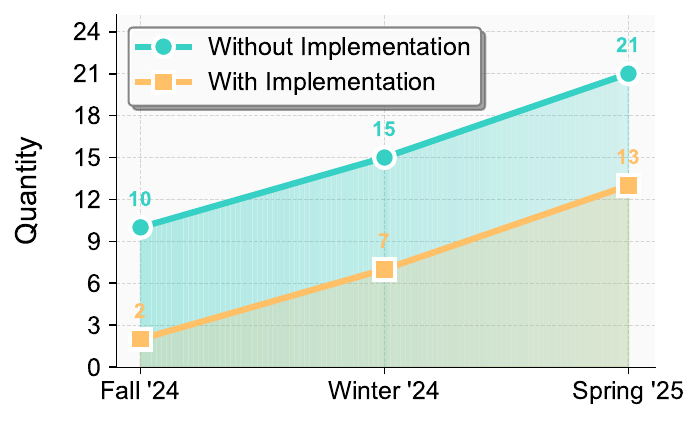} 
\vspace{-10pt}
\caption{An analysis of the number of publications in the field of AI Scientist systems on arXiv. The upper panel displays the average number of citations up to now, categorized by containing implementation details. The lower panel shows the growth in the total number of these papers with the same categorization.}
\label{fig:difference_im} 
\vspace{-10pt} 
\end{wrapfigure}

Traditionally, scientific verification requires human scientists to design, implement, and analyze experiments~\citep{popper2005logic,kuhn2014history}. In recent times, with the rapid advancement of LLMs and automated experimental tools~(e.g., AlphaFold~\citep{jumper2021highly}), the capability for AI Scientist systems to verify and falsify scientific hypotheses has seen significant development, evolving from early methods centered on tasks like direct code execution and result matching towards more sophisticated, comprehensive processes.
For instance, the \textsc{AI Scientist}~\citep{lu2024ai} designs and executes experiments with automatic code generation and error correction, enabling fully automated open-ended scientific discovery.

However, the verification and falsification process remains a significant endeavor for current AI Scientist systems. A statistical analysis of AI Scientist papers on arXiv up to May 23, 2025, reveals the challenges and community valuation related to verification and falsification. As illustrated in Figure~\ref{fig:difference_im}, 
while the overall number of publications in this domain is increasing, \textbf{studies that primarily focus on idea generation without providing concrete implementation details consistently outnumber those that incorporate such implementations.} Despite this disparity in publication volume, the upper panel of Figure~\ref{fig:difference_im} further highlights a crucial counterpoint: papers that include substantive implementation details achieve a significantly higher average number of citations, demonstrating a strong valuation within the AI Scientist community for executable verifications.

\subsection{Methods}

\begin{table}[t]
\centering
\small
\caption{State-of-the-art LLMs show relatively low accuracy on implementation benchmarks. The listed benchmarks are collected from diverse domains. The table below details their tasks, domains, scale, methods, and performance.}
\label{tab:mod_benchmarks_revised}
\resizebox{\textwidth}{!}{%
\begin{tabular}{@{} >{\centering\arraybackslash}m{6cm} >{\centering\arraybackslash}m{4cm} >{\centering\arraybackslash}m{3cm} >{\centering\arraybackslash}m{1cm} >{\centering\arraybackslash}m{3cm} >{\centering\arraybackslash}m{1.5cm} @{}}
\toprule
\textbf{Benchmark} & \textbf{Task Description} & \textbf{Domains} & \textbf{Scale} & \textbf{LLM} & \textbf{Accuracy} \\
\midrule
MLE-Bench\citep{chan2024mle} &
AI Training &
Computer Science&
75&
OpenAI o1-preview &
16.90\% \\
\rowcolor[rgb]{ .949,  .949,  .949}
CORE-Bench \citep{siegel2024core}&
Scientific Paper Reproduction &
Computer Science, Social Science, and Medicine&
270 &
OpenAI GPT-4o  &
55.56\% \\
SciReplicate-Bench \citep{xiang2025scireplicate}&
Code Generation &
Computer Science &
100&
Claude-Sonnet-3.7 &39.00\%\\
\rowcolor[rgb]{ .949,  .949,  .949}
PaperBench \citep{starace2025paperbench}&
ICML Paper Replicating&
Computer Science&
8,316&
OpenAI o1-high &
26.00\%\\
ML-Dev-Bench \citep{mldevbench}&
AI Training &
Computer Science &
30 &
Claude-Sonnet-3.5&
50.00\% \\
\bottomrule
\end{tabular}%
}
\end{table}

The ultimate goal of verification and falsification is primarily based on \textbf{designing, implementing, and analyzing experiments} to assess whether the AI-generated hypotheses are feasible. However, considering the different limitations of experiments in different scientific fields (e.g., chemistry, biology, and medicine), such as venue, equipment, and experimental safety, currently AI scientist systems focus mainly on idea verification and falsification in the field of computer science, especially in machine learning. 

For \textbf{experimental design}, the AI Scientist field has seen several innovative approaches. BioPlanner~\citep{Liu2023ReviewerGPTAE} introduces an automated evaluation framework for assessing LLMs' ability to plan experiments in the biology domain. DSBench~\citep{jing2025dsbenchfardatascience} and ScienceAgentBench~\citep{chen2024scienceagentbench} assess agents’ capabilities in experimental design for data-driven scientific discovery. 
\cite{yang2023ai} is the first work to employ an LLM as a ``masterbrain scientist'' (akin to a principal investigator), taking on the central role and steering a fully closed-loop research.
Subsequent AI Scientist systems, such as \textsc{AI Scientist}~\citep{lu2024ai} further utilize LLM agents to autonomously propose research ideas, formulate testable hypotheses, and design high-level experimental plans. These plans typically include defining variables, selecting datasets, choosing baseline models, and outlining evaluation metrics. However, the diversity and complexity of scientific experimental design present significant challenges for current AI Scientist systems. Experimental designs generated by these AI Scientist systems often lack scientific rigor, innovation, and practicality, making it difficult to meet the demands of high-level research~\citep{huang2025biomni, weng2025cycleresearcher}.

\textbf{Implementation}, theoretically the most important and difficult part of experiments, is often addressed through repository-level code generation. For instance, \textbf{LLM-based methods} such as ToolGen~\citep{Wang2024TeachingCL}, propose an approach that integrates auto-completion tools to generate repository-level code with reliable dependencies. CoCoGen~\citep{Bi2024IterativeRO} iteratively aligns and fixes errors using information extracted from the code repository. CatCoder~\citep{Pan2024EnhancingRC} enhances repository-level code generation by integrating relevant code and type context. \textbf{Agent-based methods} have showcased even more powerful performance. CodeAgent~\citep{Zhang2024CodeAgentEC} integrates five programming tools, enabling interaction with software artifacts for information retrieval, code symbol navigation, and code testing. RepoCoder~\citep{Zhang2023RepoCoderRC} streamlines the repository-level code completion process by incorporating a similarity-based retriever and a pre-trained code language model in an iterative retrieval-generation pipeline. More closely related to real-life scenarios, RepoGraph~\citep{Ouyang2024RepoGraphEA} operates at the line level, offering a more fine-grained approach compared to previous file-level browsing methods. Each node in the graph represents a line of code, and edges represent the dependencies of code definitions and references. RepoGraph boosts the success rate of existing methods by achieving an average relative improvement of 32.8\% on SWE-bench~\citep{SWEBenchCanLanguageModels}. SciCode~\citep{Tian2024SciCodeAR} decomposes the generation of scientific repository-level code into multiple sub-problems, each involving knowledge recall, reasoning, and code synthesis. AIDE~\citep{Jiang2025AIDEAE} formalizes machine learning research as a code optimization problem, and formulates trial-and-error as a tree search in the space of potential solutions. MLR-Copilot~\citep{li2024mlrcopilotautonomousmachinelearning} further proposes an ExperimentAgent to translate experiment plans into executable experimental code, crucially incorporating human feedback and iterative debugging mechanisms to manage the complexities of execution. 
In pursuit of greater autonomy, \citet{lin2025autop2c} propose AutoP2C, an LLM-based multi-agent framework that processes both textual and visual content from research papers to automatically generate executable code repositories. These ongoing developments underscore a persistent drive to overcome the inherent difficulties in scientific implementation through increasingly sophisticated methods. 

\textbf{Analysis}, as the concluding part of the experimental process, plays a crucial role in organizing experimental results into well-written experimental reports. Recent advances, such as MCX-LLM~\citep{Yen2024MCXLLMAE}, have explored the use of LLMs to convert natural language descriptions into machine-readable inputs for running Monte Carlo simulations, thus enhancing the efficiency of analysis. To evaluate the capabilities of such systems, benchmarks like FigureQA~\citep{Kahou2017FigureQAAA}, ArxivQA~\citep{Li2024MultimodalAA}, and MMSCI~\citep{li2025mmsci} have been developed, focusing on the agents' ability to understand and reason over complex figures, including graphs, charts, and tables. In addition to data interpretation, tools such as LLM-ref~\citep{Fuad2024LLMRefER} further assist researchers by supporting the synthesis of information from multiple source documents and enhancing reference management during the article writing process. Building on these advancements, state-of-the-art AI Scientist systems leverage LLM-driven code generation to automate robust statistical analyses and to generate visualizations such as plots, tables, and charts. This automation greatly streamlines the reporting of experimental findings. Furthermore, these systems often perform multiple independent experimental runs and aggregate the results, thereby enabling comprehensive meta-analyses. After the analytical stage, some AI Scientist systems~\citep{zochi2025, yamada2025ai} can even synthesize results into scientific manuscripts, automatically generating structured reports that encompass methodological details, data analyses, results, and conclusions in standard academic formats. While automatic paper writing represents a significant application of these systems, it is important to note that manuscript generation is not the core capability of AI Scientist systems. Therefore, the focus of this survey remains on the analytical and reasoning aspects rather than on the writing of final research articles.

\subsection{Evaluation}

Achieving robust verification and falsification in practice remains a profound gap for current AI Scientist systems, which is starkly illustrated by the performance of even state-of-the-art LLMs on a range of demanding benchmarks, as summarized in Table~\ref{tab:mod_benchmarks_revised}. These benchmark tasks include executing end-to-end AI model training workflows, replicating Methods and empirical results directly from scientific publications, and accurately generating executable code from complex algorithmic descriptions embedded within research papers. 
For instance, MLE-Bench~\citep{chan2024mle} assesses the ability to solve Kaggle machine learning tasks, where OpenAI's ``o1-preview'' achieved only 16.90\% medal-worthy success. PaperBench~\citep{starace2025paperbench} requires the replication of ICML research papers from scratch, on which OpenAI's ``o1-high'' manages a 26.00\% replication score. Further evidence of these challenges is found in SciReplicate-Bench \citep{xiang2025scireplicate}, a benchmark focusing on generating executable code from algorithm descriptions (Claude-Sonnet-3.7 reaches only 39.00\% execution accuracy); CORE-Bench~\citep{siegel2024core}, involving the reproduction of computational results from scientific papers across diverse fields (OpenAI's GPT-4o achieving 55.56\% accuracy on medium difficulty tasks); and ML-Dev-Bench~\citep{mldevbench}, which evaluates performance on diverse machine learning development workflow tasks (Claude-Sonnet-3.5 shows a 50.00\% success rate). These evaluations consistently demonstrate that LLMs face significant difficulty in translating conceptual understanding or initial plans into verifiably correct and operational code, highlighting a fundamental limitation in their verification capabilities and underscoring the importance of systematic verification and implementation capabilities for the maturation of AI Scientist systems.

%% file: sub_tex/review.tex
\begin{simpleElegantQuote}
\textbf{Definition:} An AI reviewer system comprehensively assesses the generated manuscripts to provide insightful reviews, aiming to enhance the quality of the scientific artifact produced by AI Scientist systems. 
\end{simpleElegantQuote}
Peer review is essential for refining scientific work, but traditional systems face significant challenges especially in fast-evolving fields like computer science, where new results can become outdated before they are published. At the same time, the peer review system depends on a limited number of reviewers who are often overwhelmed by the high volume of submissions, which leads to delays, biases, inconsistencies, and sometimes unfair outcomes~\citep{Stelmakh2020PriorAP, Zhang2022InvestigatingFD, c1f07b2bae5a41d1bf0278eaa7fa158b}. To address these problems, researchers are actively exploring the use of AI systems to support scientific reviewing throughout the peer review life-cycle. Broadly speaking, the development of AI reviewer systems can be divided into two phases, marked by the emergence of LLMs: \textbf{(1) Pre-LLM phase: }Earlier AI reviewer systems primarily focused on automating reviewer-paper matching based on content analysis and reviewer expertise~\citep{charlin2013toronto,stelmakh2019peerreview4all,stelmakh2021towards}. These systems also handle screening and pre-check processes, such as verifying compliance with submission guidelines, plagiarism, and anonymity checks~\citep{Goldberg2024UsefulnessOL, lee-etal-2025-plagbench}. 
\textbf{(2) LLM-based AI reviewer system: }With the advent of LLMs, the capabilities of AI reviewers have expanded dramatically. A key advancement is the ability to automatically generate detailed, structured reviews. These reviews typically include summaries, assessments of strengths and weaknesses, and constructive feedback, which aim to make the review process more efficient and equitable, ultimately enhancing both the speed and quality of scientific evaluation~\citep{wang2020reviewrobot,idahl2024openreviewer,tyser2024openreviewer,pendyala2025automated, zhu2025deepreview}.

\subsection{Methods}
\label{subsec:methodologies_frameworks_report}

\begin{figure}[t]
    \centering
    \includegraphics[width=\linewidth]{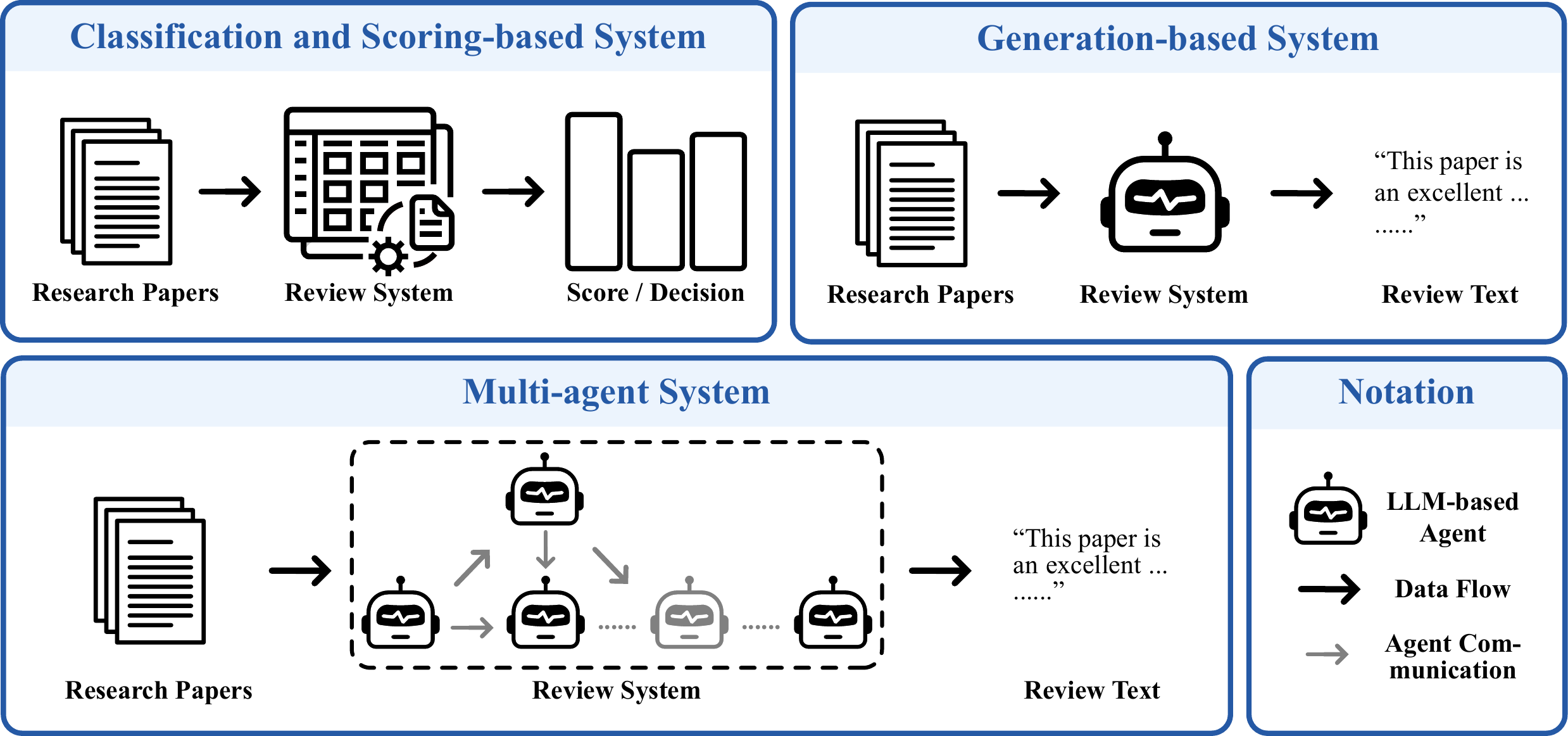}
    \caption{Three paradigms of AI reviewer systems with increasing complexity. The process begins with (1) classification \& scoring systems that provide quantitative outputs (e.g., scores or accept/reject decisions). This paradigm gradually evolves into (2) generation Systems that produce narrative review text. Recently, a more advanced paradigm employs (3) multi-agent Systems, where multiple AI agents collaborate to create a comprehensive, multi-faceted evaluation.}
    \label{fig:review_method}
\end{figure}

Different technologies for AI reviewer systems have been developed to handle specific parts of the review process. These methods range from simple tasks, such as screening and classifying submissions~\citep{Wang2024ZeroshotGL, Cao2024PromptingIA, Jaumann2025LGARZL}, to more complex tasks, including writing review comments, simulating reviewer discussions, and reasoning about a paper’s content~\citep{Peng2024ReviewLLMHL,Su2025ReviewriterAI, Sun2024MetaWriterET, Zhou2025LargeLM,zhu2025deepreview, jin-etal-2024-agentreview}. We classify these methods based on their underlying paradigms and the complexity of the tasks they address~(Figure~\ref{fig:review_method}).

\textbf{Classification- and scoring-based AI reviewer systems} primarily focus on evaluating academic papers to assign quantitative scores against predefined criteria~\citep{zhu2025deepreview}, classifying them based on specific attributes (e.g., relevance to a conference~\citep{leyton2024matching}), or screening literature for inclusion in systematic reviews \citep{joos2024cutting}. These systems often target specific, measurable aspects of an academic paper rather than generating holistic reviews~\citep{wang2024autosurvey,liang2025surveyx}. For example, \textbf{ReviewRobot}~\citep{wang2020reviewrobot} employs a knowledge graph-based framework to predict review scores and generate structured evidence, emphasizing explainability. Its performance in score prediction reaches 71.4\% accuracy, with a significant portion of its generated comments deemed valid and constructive by human experts. Similarly, \textbf{RelevAI-Reviewer} \citep{couto2024relevai} conceptualizes survey paper review as a classification task, benchmarking AI's ability to assess relevance using a dataset of over 25,000 instances. In terms of empirical experiment, the NeurIPS 2024 committee employed LLMs to systematically categorize submitted papers and check them against a standardized checklist \citep{goldberg2024usefulness}. Over 70\% of authors found the system useful, and 70\% made substantial revisions based on the detailed feedback.

\textbf{Generation-based AI reviewer systems} are designed to produce academic reviews in natural language, mirroring the narrative outputs of human reviewers. For example, 
\textbf{Reviewer2} \citep{gao2024reviewer2} introduces a two-stage framework that first models the distribution of potential review aspects and then generates prompts to guide an LLM in producing detailed academic reviews. \textbf{OpenReviewer} \citep{tyser2024openreviewer}, a specialized 8-billion parameter LLM fine-tuned on 79,000 expert reviews, processes PDF submissions to generate structured reviews adhering to conference guidelines, notably producing more ``critical and realistic reviews'' than general-purpose LLMs like GPT-4. Another automated method described in \citep{wu2024automated} uses LLMs to analyze papers, extract key information, and generate insights with quality control measures to mitigate hallucination, reportedly matching manual review quality in a case study. Furthermore, \textbf{CycleReviewer}~\citep{weng2025cycleresearcher} provides a suite of specially trained LLMs to generate expert-level opinions and evaluation scores, achieving a 26.89\% reduction in MAE for score prediction compared to individual human reviewers. The critical challenge for these systems lies in moving beyond fluent text generation to provide genuinely insightful and constructive criticism, avoiding the tendency towards overly positive or superficial assessments~\citep{Shojaee2025TheIO, Laskar2024ASS}.

\begin{table*}[tbp]
\centering
\footnotesize
\caption{An overview of AI Reviewer benchmarks, including task focus, sample scale, and data source. }
\label{tab:ai_reviewer_datasets_benchmarks_revised}
\begin{tabular}{@{}>{\centering\arraybackslash}m{5.2cm} >{\centering\arraybackslash}m{3cm} >{\centering\arraybackslash}m{1.2cm} >{\centering\arraybackslash}m{4.0cm}@{}}
\toprule
\textbf{Name} & \textbf{Task Focus} & \textbf{Scale} & \textbf{Source} \\
\midrule

PeerRead \citep{kang2018dataset} & Predictive Modeling & 14700 & ACL, NeurIPS, ICLR \\
\rowcolor[rgb]{ .949,  .949,  .949}
ASAP-Review \citep{yuan2022can} & Prediction \& Generation & 8877 & ICLR, NeurIPS\\
MOPRD \citep{lin2023moprd} & Review Generation & 6578 & Journals from PeerJ\\
\rowcolor[rgb]{ .949,  .949,  .949}
RelevAI-Reviewer \citep{couto2024relevai} & Predictive Modeling & 25164 & Existing Survey Papers \\
ReviewMT \citep{tan2024peer} & Review Generation & 26841 & ICLR, Nature Communications \\
\rowcolor[rgb]{ .949,  .949,  .949}
REVIEWER2 \citep{gao2024reviewer2} & Review Generation & 27000 & Computer Science Conferences\\
{SWIF$^{2}$T} \citep{chamoun2024automated} & Review Generation & 300 & Curated Peer Reviews \\
\rowcolor[rgb]{ .949,  .949,  .949}
Automatic Evaluation Metrics \citep{hopner2025automatic} & Predictive Modeling & 15002 & OpenReview, Semantic Scholar\\
PeerQA \citep{baumgartner2025peerqa} & Text Analysis & 579 & Papers from computer science, Geoscience, Health\\
\rowcolor[rgb]{ .949,  .949,  .949}
AI/Human Peer Review \citep{yu2025your} & Text Analysis & 788984 & ICLR, NeurIPS\\
Review-5K \citep{weng2025cycleresearcher} & Prediction \& Generation & 24991 & ICLR 2024\\
\rowcolor[rgb]{ .949,  .949,  .949}
DeepReview \citep{zhu2025deepreview} & Review Generation & 14664 & ICLR 2024, 2025\\
\bottomrule
\end{tabular}
\end{table*}

\textbf{Multi-agent AI reviewer systems.} 
Given that peer review is inherently a collaborative process~\citep{miyao2019does, beygelzimer2023has}, researchers have begun exploring AI reviewer systems based on multi-agent frameworks. These systems typically employ multiple interacting AI agents with distinct roles (e.g., meta-reviewer, area chair) to simulate discussions, thereby producing more comprehensive and nuanced reviews. For example, AgentReview~\citep{jin-etal-2024-agentreview} uses LLM agents to investigate review dynamics and latent biases; ReviewAgents~\citep{gao2025reviewagents} proposes a multi-role framework to emulate human reasoning processes; and MARG~\citep{d2024marg} distributes the paper among specialized agents to improve feedback quality. 
Taking this collaborative structure further, the current frontier of such systems is to replicate deeper human cognitive processes through advanced reasoning and emulation. This involves incorporating multi-stage reasoning, structured analytical frameworks, and evidence-based argumentation to approximate the critical depth achieved by human experts. A prime example is the DeepReviewer framework~\citep{zhu2025deepreview}, which uses multi-stage thinking processes (Analysis, Argument, Assessment) and has demonstrated strong performance and resilience~\citep{ye2024we}. Similarly, \textsc{AI Scientist}~\citep{lu2024ai} employs multi-round reflections to evaluate papers. By combining multi-agent structures with advanced cognitive emulation, these systems are moving toward a ``glass box'' approach in which the AI's deliberative process is more discernible and verifiable, serving as an essential step for building trust in high-stakes academic assessment.

\subsection{Evaluation Benchmarks}
\label{subsec:review_datasets_benchmarks}
The advancement and rigorous assessment of AI reviewer systems fundamentally rely on the availability of appropriate datasets and benchmarks. These resources \citep{choudhary2021react,li2022peersum,guo-etal-2023-automatic,purkayastha-etal-2023-exploring,bharti2024politepeer,darcy-etal-2024-aries,zhou-etal-2024-llm,lou2025aaar10assessingaispotential} serve as both the training ground for evaluation models and the standard against which their capabilities are measured \citep{zhuang2025large}. Early datasets are often collected from conference or journal review records~\citep{Lin2022MOPRDAM, Zeng2023ScientificOS, Zhou2024IsLA}. As the field grows, researchers begin creating more specialized resources to reflect the different tasks involved in peer review. This includes resources specifically curated for assessing relevance determination \citep{couto2024relevai}, question-answering abilities \citep{baumgartner2025peerqa}, the generation of aspect-guided \citep{gao2024reviewer2} or weakness-focused feedback \citep{chamoun2024automated}, and even the crucial task of detecting AI-generated reviews \citep{yu2025your,weng2025cycleresearcher}. 
Table~\ref{tab:ai_reviewer_datasets_benchmarks_revised} summarizes information on key datasets and benchmarks for AI Reviewer Evaluation, providing a snapshot of the resources available.

\subsection{Current AI Scientist Systems Are Not Good Enough}
\label{subsec:llm_judge_assessment}

\begin{table}[t]
\centering
\small
\caption{
Evaluation of AI-generated papers produced by various AI scientist systems. Scores represent the average ratings given by DeepReviewer-14B~\citep{zhu2025deepreview} across the number (``Num'') of available papers. The ``Percentile'' column shows each system’s relative quality ranking. Note: Publicly available papers may be curated and therefore may not fully represent the typical output of each system.
}
\label{tab:ai_scientist_performance_llm_judge}
\begin{adjustbox}{width=\textwidth}
\begin{tabular}{ccccccc}
\toprule
\textbf{AI Scientist Systems} & \textbf{Number}  & \textbf{Soundness} & \textbf{Presentation} & \textbf{Contribution} & \textbf{Rating}& \textbf{Percentile}\\
\midrule
\textsc{AI Scientist}~\citep{lu2024ai} & 10  & 2.08 & 1.80 & 1.75 & 3.35&8.22\% \\
\rowcolor[rgb]{ .949,  .949,  .949}
HKUSD AI Researcher~\citep{airesearcher} & 7 & 1.75 & 1.46 & 1.57 & 2.57 &3.43\%\\
\textsc{AI Scientist-v2}~\citep{yamada2025ai} & 3 & 1.67 & 1.50 & 1.50 & 2.33 &2.04\% \\
\rowcolor[rgb]{ .949,  .949,  .949}
CycleResearcher-12B~\citep{weng2025cycleresearcher} & 6  & 2.25 & 1.75 & 2.13 & 3.75& 16.88\%\\
Zochi~\citep{zochi2025} & 2  & 2.38 & 2.38 & 2.25 & 4.63&29.96\% \\
\bottomrule
\end{tabular}
\end{adjustbox}
\end{table}

\setlength{\intextsep}{5pt}
\setlength{\columnseprule}{10pt}
\begin{wraptable}{r}{0.46\textwidth}
  \centering
  \caption{Twelve major defect categories detected in the 28 assessed papers. }
  \label{tab:defect_categories}
  \small
\resizebox{0.45\textwidth}{!}{
  \begin{tabular}{c c c}
    \toprule
    \textbf{Defect Category} & \textbf{Number} & \textbf{Percentage} \\
    \midrule
    Experimental Weakness & 28 & 100\% \\
    \rowcolor[rgb]{ .949,  .949,  .949}
    Methodological Unclarity/Flaws & 27 & 96.4\% \\
    Writing \& Presentation Issues & 26 & 92.9\% \\
    \rowcolor[rgb]{ .949,  .949,  .949}
    Novelty Concerns & 25 & 89.3\% \\
    Theoretical Weakness & 24 & 85.7\% \\
    \rowcolor[rgb]{ .949,  .949,  .949}
    Literature Review Deficiencies & 22 & 78.6\% \\
    Practicality \& Robustness Gaps & 21 & 75.0\% \\
    \rowcolor[rgb]{ .949,  .949,  .949}
    Reproducibility Issues & 20 & 71.4\% \\
    Computational Cost Concerns & 18 & 64.3\% \\
    \rowcolor[rgb]{ .949,  .949,  .949}
    Component Analysis & 16 & 57.1\% \\
    Hyperparameter Analysis Lacking & 16 & 57.1\% \\
    \rowcolor[rgb]{ .949,  .949,  .949}
    Ethical Considerations Missing & 3 & 10.7\% \\
    \bottomrule
  \end{tabular}}
\end{wraptable}

While pioneering AI Scientist systems~\citep{yamada2025ai, zochi2025} demonstrate capabilities such as generating workshop-accepted manuscripts, we conduct a rigorous evaluation that reveals their persistent deficiencies in scientific rigor. To quantify these gaps, we employ DeepReviewer-14B~\citep{zhu2025deepreview}, an advanced AI reviewer model, to assess 28 publicly available research papers produced by 5 leading AI Scientist systems. Though public availability may bias toward higher-quality outputs, this analysis exposes systemic limitations in current autonomous research.
The evaluation results are summarized in Table~\ref{tab:ai_scientist_performance_llm_judge}, painting a concerning picture: the highest-rated system achieves an average rating of only 4.63 out of 10, with most systems scoring considerably lower. These low scores reflect poor performance across key metrics, including soundness, presentation, and contribution. Furthermore, the evaluation results identify twelve major defect categories present in the assessed papers~(Table~\ref{tab:defect_categories}), with \textbf{``Experimental Weakness'' appearing in 100\%} of the papers. This universal deficiency highlights severe limitations in the current implementation capabilities of AI scientist systems, particularly regarding experimental design, execution, and result analysis.

Other prevalent issues includes ``Methodological Unclarity/Flaws'' (96.4\%), ``Writing \& Presentation Issues'' (92.9\%), and ``Novelty Concerns'' (89.3\%). These findings suggest that current AI Scientist systems not only struggle with scientific execution but also stuck with clearly articulating their research findings. The high incidence of ``Theoretical Weakness'' (85.7\%), and ``Literature Review Deficiencies'' (78.6\%) further indicates that these systems often fail to offer genuinely original contributions or ground their claims in robust theoretical frameworks. 
These evaluation results align with the low quantitative scores presented in Table~\ref{tab:ai_scientist_performance_llm_judge}, clearly demonstrating that current AI scientist systems, despite their technological sophistication,\textbf{ cannot independently produce scientific artifacts that meet established standards for high-quality scientific communication.} 
These findings strongly suggest that current AI-generated research often lacks the depth, rigor, and implementation quality expected in meaningful autonomous research and credible scientific contributions. Addressing these limitations through \textbf{evolutionary mechanisms becomes necessary} for advancing AI Scientist systems toward their full potential as autonomous scientific agents.

%% file: sub_tex/evolution.tex
\begin{simpleElegantQuote}
\textbf{Definition:} Evolution is the capability to continuously \textbf{advance overall research abilities based on feedback from internal reflection or external inputs.} This capability involves the dynamic planning for research directions and the autonomous learning for improvement, serving as the pathway for an elementary AI Scientist to evolve into a mature scientific agent.
\end{simpleElegantQuote}

Several influential works have already emerged in the field of AI Scientists~\citep{lu2024ai,weng2025cycleresearcher,zochi2025, yamada2025ai}. For instance, \textsc{AI Scientist-v2}~\citep{yamada2025ai} is capable of autonomously generating manuscripts that successfully pass peer review at workshops of major machine learning conferences. Despite this substantial progress, extensive quantitative evidence~(Tables~\ref{tab:mod_benchmarks_revised}, \ref{tab:ai_scientist_performance_llm_judge}, \ref{tab:defect_categories}) indicates that systemic issues remain in the scientific rigor and implementation quality of current AI-generated research, and \textbf{there is still a considerable gap before AI Scientists are able to make ground-breaking discoveries that solve grand challenges} across medicine, energy, and the environment, thereby changing the world and reshaping the scientific research paradigm. We attribute this gap to (1) \textbf{the inherent limitations of the foundation models} (i.e., LLMs) that underlie AI Scientists, and (2) \textbf{the inadequate scientific research abilities} of current AI Scientists, such as generating hypotheses with low feasibility, employing incorrect validation methods, and having limited ability to understand and decompose complex research tasks. A more detailed discussion of this gap is deferred to Section~\ref{sec:discussion}. 

Through continuous evolution, an elementary AI Scientist can progressively bridge this gap and gradually approach the capabilities of a mature scientific agent. Specifically, evolutionary mechanisms enable AI Scientists to transcend the static confines of their initial foundation models and refine their research abilities through iterative feedback and interaction with both human experts and dynamic scientific environments, thereby advancing toward the long-term vision of autonomous, reliable, and innovative scientific discovery. In this section, we focus on discussing the key technologies required for evolution capability, including strategies for \textbf{dynamic planning} of research directions and methods for \textbf{autonomous learning}~\citep{schmidhuber2007godel}. These foundational technologies are essential for closing the gap that still separates current AI Scientist systems from making transformative discoveries.

\subsection{Methods}
\label{}

Currently, most research in the field of AI Scientist focuses on the evolution of individual scientific artifacts (such as a scientific hypothesis or an experimental code implementation), rather than on managing long-term research cycles with comprehensive planning and iteration. In this section, we primarily summarize how existing AI Scientist systems approach evolution by reviewing their approaches to planning evolution paths~(Section~\ref{subsec:planning}) and employing autonomous improvement~(Section~\ref{subsec:iterative}). The discussion of long-cycle research iteration and comprehensive planning is deferred to Section~\ref{sec:future_and_conclusion}.

\subsubsection{Dynamic Planning}
\label{subsec:planning}

The dynamic planning capability aims to explore the most valuable research directions within a constrained search space, enabling an AI Scientist to efficiently allocate resources, prioritize hypotheses, and adaptively refine research trajectories based on ongoing results and feedback. \textbf{Currently, in the field of AI Scientist, dynamic planning remains relatively underdeveloped}, leaving a few cutting-edge studies employing tree-search strategies with LLMs to enable structured exploration of diverse scientific hypotheses~\citep{zochi2025} and detailed experimental plans~\citep{jansen2024discoveryworld, yamada2025ai, yuan2025dolphin}. For instance, when given a broad research domain, Zochi~\citep{zochi2025} conducts a comprehensive process of exploration and refinement process, where it generates multiple candidate hypotheses, designs experiments to test them, and iteratively improves its strategy based on the results. In \textsc{AI Scientist-v2}, ~\citet{yamada2025ai} introduce an experiment manager agent in combination with a novel agentic tree-search algorithm to generate and refine code implementations. Subsequent experiments leverage the top-performing code checkpoints from the tree search to iteratively evaluate various research hypotheses. Although these approaches enable the discovery of higher-quality scientific hypotheses and experimental designs, they still face challenges such as increased complexity, greater computational cost, and limited scalability~\citep{jansen2025codescientist}.

\subsubsection{Autonomous Learning}
\label{subsec:iterative}
Various autonomous learning strategies have been employed to continuously enhance the overall research capabilities of the AI Scientist. These approaches can be categorized into self-reflection methods and externally guided methods, depending on the source of feedback.

\textbf{Self-reflection methods}~\citep{lu2024ai, romera2024mathematical,weng2025cycleresearcher, jansen2025codescientist, yamada2025ai, novikov2025alphaevolve} refer to the process where the LLM serves as its own feedback provider, iteratively evaluating and refining its outputs until a certain quality standard is met~\citep{madaan2023self, pan2024automatically}. This concept of continuous self-improvement has been widely adopted in LLMs to enhance downstream performance~\citep{weng2022large,zelikman2022star} and reduce harmful responses~\citep{bai2022constitutional}. 
Accordingly, in the field of AI Scientist, researchers have also begun employing self-reflection strategies to \textbf{improve the quality of generated scientific artifacts.} For example, \citet{weng2025cycleresearcher} propose an iterative preference training framework that consists of two components: (1) CycleResearcher, which performs research tasks, and (2) CycleReviewer, which simulates the peer review process by providing iterative feedback through reinforcement learning. This entire procedure is refined iteratively, leading to progressively improved research capabilities with each cycle. DeepMind’s FunSearch~\citep{romera2024mathematical} follows a similar evolutionary paradigm, pairing a pretrained LLM, tasked with generating creative solutions to scientific problems, with a systematic evaluator that guards against confabulations and incorrect reasoning. It iteratively samples the best-performing code programs and incorporates them into new prompts for the LLM to build upon, evolving initially low-scoring programs into high-performing ones and thereby uncovering novel insights. 

\textbf{Externally guided methods}~\citep{yuan2025dolphin,yu2024researchtown, schmidgall2025agentrxiv, jansen2024discoveryworld}. External evaluation of the AI-generated artifacts also serves as an important form of feedback. Current AI Scientist systems generally adopt two approaches: (1) incorporating human supervision during the generation process to assess the quality of hypotheses and provide brief revision suggestions~\citep{jansen2025codescientist,zochi2025}; (2) leveraging internet-based academic platforms (e.g., Semantic Scholar~\citep{fricke2018semantic}) to filter out hypotheses that are overly similar to existing literature, thereby ensuring novelty~\citep{yuan2025dolphin, lu2024ai, weng2025cycleresearcher}. Beyond these methods, some efforts aim to build research communities composed entirely of AI Scientist systems, enhancing the quality of their outputs through collaboration. For example, \citet{yu2024researchtown} introduce RESEARCHTOWN, a multi-agent framework for simulating research communities. RESEARCHTOWN models the research community as an ``agent-data graph'', where AI researchers and papers are represented as nodes, and activities such as reading, writing, and reviewing are implemented as message-passing operations within a TextGNN inference framework. Similarly, \citet{airaXiv2025} establish AiraXiv, a centralized platform that archives scientific artifacts generated by AI Scientist systems, allowing them to collaborate, share insights, and iteratively build upon each other’s work. However, \textbf{the field currently lacks a specialized protocol for scientist-to-scientist communication} to enable more efficient and structured interactions among AI scientist systems. Such a protocol could standardize interaction workflows and define communication formats, ultimately facilitating effective collaboration and information exchange between AI Scientist systems. Hence, developing a specialized communication protocol for AI Scientist systems represents a key future direction for advancing automatic scientific discoveries (see Section~\ref{sec:future_and_conclusion} for further discussion).

%% file: sub_tex/discussion.tex
\begin{table*}[t]
    \centering
    \large
    \caption{Comparison of survey papers on LLM-based scientific agents, scientific tools, and AI Scientist. \textbf{Research Scope:} Domains of each survey paper; \textbf{Ethics/Society:} discussion of ethical and societal issues; \textbf{Capability Roadmap:} whether a staged roadmap toward a mature AI Scientist is provided; \textbf{Critical Gap Assessment:} whether the paper critically evaluates progress toward the mature AI Scientist and identifies bottlenecks. \CheckmarkBold: strong/unique coverage; $\triangle$: partial/related coverage; \XSolidBrush: little or no coverage.}
    \label{tab:ai_scientist_survey_comparison}
    \renewcommand{\arraystretch}{1.25}
    \resizebox{0.98\textwidth}{!}{
    \begin{tabular}{cccccccc}
        \toprule
        \textbf{Survey} & \textbf{Research Scope} & \textbf{Ethics/Society} & \textbf{Capability Roadmap} & \textbf{Critical Gap Assessment} \\
        \midrule
        \cite{zhang2024comprehensive}    & Scientific LLMs & \XSolidBrush & \XSolidBrush & \XSolidBrush \\
        \rowcolor[rgb]{ .949,  .949,  .949}
       \cite{ren2025towards}     & LLM-based Scientific Agents & \CheckmarkBold & \XSolidBrush & \XSolidBrush \\
        \cite{gridach2025agentic}  & Agentic AI for scientific discovery & \CheckmarkBold & \XSolidBrush & \XSolidBrush \\
        \rowcolor[rgb]{ .949,  .949,  .949}
        \cite{zheng2025automation}    & LLMs in science & $\triangle$ & \CheckmarkBold & \XSolidBrush \\
        \cite{zhou2025hypothesis}     & AI-driven research support systems & \XSolidBrush & $\triangle$ & \XSolidBrush \\
        \rowcolor[rgb]{ .949,  .949,  .949}
        \cite{luo2025llm4sr}      & LLMs for scientific research  & \XSolidBrush & $\triangle$ & \XSolidBrush \\
         \cite{Goyal2025survey}           & AI Scientist  & \CheckmarkBold & \XSolidBrush & \XSolidBrush \\
        
        \midrule
        \rowcolor[rgb]{ .949,  .949,  .949}
        \textbf{Ours}     & AI Scientist  & \CheckmarkBold & \CheckmarkBold & \CheckmarkBold \\
        \bottomrule
        \end{tabular}}
    
\end{table*}

\subsection{Positioning of This Survey}
Although previous surveys have provided valuable overviews of LLM architectures, agentic frameworks, and scientific automation tools for automatic scientific discovery~\citep{zhang2024comprehensive, zheng2025automation, ren2025towards, zhou2025hypothesis, Goyal2025survey, luo2025llm4sr}, there remains a critical gap: \textbf{a capability-based roadmap that comprehensively charts the path toward the ideal AI scientist}. Our survey addresses this gap by introducing a capability-level framework~(Figure~\ref{fig:capability-level}) that systematically defines the stages of AI scientist development, establishing clear benchmarks for the next era of AI-driven scientific discovery. We critically analyze current research through the lens of this capability framework, identifying key bottlenecks and missing components necessary for the emergence of ideal scientific agents. The difference between our survey and existing survey papers is shown in Table~\ref{tab:ai_scientist_survey_comparison}.

\subsection{
Limitations of Current AI Scientists in Scientific Discovery
}
\label{subsec:limitation}

Despite the substantial progress in the field of AI Scientist systems, 
there is still a considerable gap before AI Scientist can mature into scientific agents capable of making ground-breaking discoveries, thus changing the world and reshaping the scientific research paradigm. This persistent gap can be attributed to several critical factors. First, \textbf{the inherent limitations of the foundation models} (i.e., LLMs) that underpin AI Scientist systems pose significant barriers. These models, while demonstrating impressive capabilities in language understanding and generation, often lack the depth of domain expertise, capacity for scientific reasoning, and nuanced judgment required for high-impact scientific breakthroughs. \textbf{Their outputs are constrained by the knowledge encoded in their training data}, which may be outdated, incomplete, or not fully aligned with the latest advancements in the scientific community. Second, current AI Scientist systems \textbf{exhibit inadequate scientific research abilities} in several key aspects. For instance, they may generate hypotheses that lack feasibility or novelty, apply inappropriate or insufficient validation methods, and struggle with the deep comprehension and decomposition of complex research problems. These shortcomings not only limit the reliability of AI-generated scientific artifacts but also hinder the AI Scientist systems' abilities to independently advance the frontiers of science.

\subsubsection{Fundamental Limitations of Base Models}
Despite their remarkable progress, LLMs that serve as the foundation for AI Scientist systems exhibit several intrinsic limitations, including hallucinations, high costs and inefficiencies in knowledge updating, and catastrophic forgetting. These challenges critically constrain the scientific utility and reliability of current AI Scientist systems, making it crucial to address them for the advancement of AI-driven scientific research.

\textbf{Hallucination.} In scientific research, precision, factual accuracy, and reliability are paramount. However, LLMs are prone to ``hallucination'', a phenomenon where the model generates plausible-sounding but false or unverifiable content~\citep{huang2025survey, zhang2025siren}. This can manifest in several ways: (1) The model may fabricate research findings, experimental data, references, or even entirely non-existent theories, presenting them as factually correct. (2) AI-generated citations, data, and formulas are often unreliable or improperly sourced, making it difficult for human researchers to trace and verify the origin and validity of the information. (3) Hallucinated outputs may contain subtle logical inconsistencies or contextually inappropriate claims, which can mislead researchers, compromise the credibility of scientific outputs, and ultimately undermine trust in AI-generated scientific artifacts. Addressing hallucination remains one of the most significant challenges for deploying LLM-based AI Scientist systems in high-stakes scientific contexts.

\textbf{High cost and inefficiency of knowledge updating.}
Cutting-edge scientific research advances rapidly, requiring AI Scientist systems to constantly incorporate the latest findings and data. However, keeping LLMs up-to-date is non-trivial: (1) The volume of new scientific literature and data is immense, and integrating this information into an LLM demands significant computational resources, storage, and time. (2) Frequent retraining or fine-tuning of large-scale models is both costly and technically challenging, often leading to delays in knowledge integration that may render the model outdated in fast-moving research domains. (3) The lack of efficient, fine-grained, and continuous learning mechanisms means that knowledge refreshing processes are typically slow, creating a persistent lag between the state of the art in scientific fields and the knowledge encoded in AI Scientist systems.

\textbf{Catastrophic Forgetting.}
Another fundamental challenge is catastrophic forgetting, which is the tendency of neural networks to lose previously acquired information when trained on new data~\citep{kirkpatrick2017overcoming}. For an AI Scientist, this challenge means that attempts to update the model with new scientific knowledge may inadvertently overwrite or degrade its understanding of existing scientific concepts, methods, and facts. Such instability undermines the system’s ability to maintain a comprehensive and coherent body of scientific knowledge over time. As a result, integrating findings across different periods and scientific fields becomes particularly challenging, which in turn impairs the LLM-based AI Scientist systems' capacities to perform longitudinal analyses, connect historical insights with emerging discoveries, and deliver consistent, reliable support for complex scientific explorations.

\subsubsection{Limitations of Research Capabilities of AI Scientist Systems}
The research capabilities of current AI Scientist systems remain limited in several crucial aspects. These limitations not only restrict their effectiveness in supporting and automating scientific research but also constrain their potential to independently drive significant scientific breakthroughs. Below, we discuss core capability gaps through the lens of the proposed capability framework~(Figure~\ref{fig:capability-level}).

\textbf{Knowledge acquisition.}
The ability to acquire, organize, and internalize knowledge from the vast and rapidly expanding body of scientific literature is the fundamental capability for an AI Scientist. This includes not only searching for and retrieving relevant papers and results, but also extracting core ideas, incorporating experimental details, and integrating disparate findings into a coherent understanding. However, existing AI Scientist systems face several substantial challenges in this area, including: \textbf{(1) Precision in information retrieval.} The enormous scale and diversity of scientific literature make it difficult for current AI Scientist systems to accurately retrieve the most relevant and up-to-date information for a given research problem~\citep{landhuis2016scientific}. They often struggle with the precise interpretation of complex or ambiguous queries, effective filtering of highly relevant documents, and ensuring that the results remain timely and reliable. Both precision and recall in retrieval must be further improved to meet scientific standards. \textbf{(2) Multi-document summarization and synthesis.} Generating accurate and comprehensive literature reviews requires not only retrieving multiple relevant documents, but also synthesizing information across sources, resolving inconsistencies, and generating faithful summaries~\citep{agarwal2024litllm}. Recent approaches, such as RAG and plan-based summarization, have made progress in this area. However, studies have shown that LLMs are still prone to misquoting or distorting source materials~\citep{agarwal2024llms}, indicating persistent challenges in semantic understanding, information integration, and factual accuracy when summarizing scientific knowledge from multiple documents.

\textbf{Idea generation.} One of the most critical aspects of scientific research is the generation of novel and feasible scientific hypotheses. Despite notable advancements, the idea generation capability remains significantly constrained by several limitations: \textbf{(1) Difficulty in generating high-quality hypotheses.} Generating scientific hypotheses that are not only novel but also feasible is a complex task that requires a deep understanding of both existing knowledge and the ability to identify innovative connections. However, empirical evidence suggests that current AI Scientist systems often fall short in this regard. Specifically, the ideas produced by these models tend to lack true originality and are frequently repetitive across different runs or even across different models~\citep{lu2024ai}. The underlying cause of this phenomenon is twofold: First, LLMs are fundamentally limited by their training data, which constrains their ability to move beyond well-trodden conceptual ground. Second, without advanced mechanisms for knowledge integration or creativity, AI Scientist systems tend to rely heavily on superficial text correlations, rather than generating ideas with genuine scientific novelty. This not only leads to repetitive outputs but also limits the potential for breakthrough discoveries. \textbf{(2) Challenges in evaluating the quality of AI-generated scientific hypotheses.} A further limitation arises from the challenge of reliably assessing the quality and potential impact of AI-generated scientific hypotheses. Unlike standardized tasks with clear evaluation criteria, the value of a scientific hypothesis is often subjective, context-dependent, and difficult to quantify. Current evaluation methods rely heavily on human expert judgment, which is not only resource-intensive but also susceptible to subjectivity and inconsistency. Recent attempts to introduce automated evaluators have helped to some extent, but these systems themselves inherit the biases and knowledge limitations of their underlying models. As a result, the lack of reliable, objective, and scalable evaluation frameworks for AI-generated hypotheses remains a bottleneck for the advancement of AI Scientist systems in real-world research settings.

\textbf{Verification and falsification.} As discussed in Section~\ref{sec:verification_and_falsification}, verification and falsification is a complex capability that requires rigorous verification and potential falsification of hypotheses through experiment design, execution, and validation. For AI Scientist systems, this process is both technically demanding and operationally complex, especially when considering the integration of multi-agent frameworks and collaboration with external tools or human researchers: \textbf{(1) Multi-agent collaboration. }
In modern research environments, effective collaboration, whether with other AI agents, human scientists, or a variety of external tools, is essential for scientific progress~\citep{guo2024large,qian2024scaling,pu2025piflowprincipleawarescientificdiscovery}. An AI Scientist should not only understand collaborative protocols and division of labor but also reliably coordinate with diverse stakeholders, execute assigned tasks, and integrate outputs back into larger research workflows~\citep{bo2024reflective,zhang2024chain}. Current AI Scientist systems, however, still exhibit significant weaknesses in robustness, adaptability, and fault tolerance, particularly in dynamic or unpredictable environments~\citep{wei2025browsecomp}. For instance, they frequently encounter difficulties when interacting with changing external APIs, handling error messages, or managing concurrent tasks, which are all crucial for real-world scientific automation~\citep{shen2025shortcutsbenchlargescalerealworldbenchmark}. \textbf{(2) Experiment design, execution and validation.}
The core of the verification and falsification capability lies in experiment design, execution, and validation. However, current AI Scientist systems face multiple challenges across these stages. Firstly, their automatically generated experimental designs often lack scientific rigor, innovation, and practicality, making it difficult to satisfy the requirements of advanced research. Additionally, these systems tend to rely on static templates and a narrow set of literature, which limits their ability to incorporate the latest technological advances and experimental methodologies. Consequently, the resulting experimental plans are frequently disconnected from the research frontier and lack real-world applicability~\citep{weng2025cycleresearcher, zochi2025}. Furthermore, as shown by the empirical evidence in Table~\ref{tab:mod_benchmarks_revised}, an AI Scientist encounters significant difficulties when translating conceptual understanding or initial plans into verifiably correct and executable code. For instance, on state-of-the-art benchmarks such as SciReplicate-Bench~\citep{xiang2025scireplicate}, the best execution accuracy achieved is only \textbf{39\%}, which highlights a fundamental limitation in their execution and validation capabilities.

\textbf{Evolution.} The ability of an AI Scientist to continuously evolve is essential for sustained scientific innovation. 
However, current AI Scientist systems demonstrate notable limitations in their evolutionary capabilities, which hinder their long-term progress and adaptability: \textbf{(1) Dynamic planning} refers to the system's ability to adaptively explore new research directions, update hypotheses, and iteratively refine experimental strategies. Despite its importance, this capability remains underdeveloped in existing AI Scientist systems. Recent studies have introduced tree-search-based strategies. While these represent a meaningful step forward, they are still constrained by several factors: increased algorithmic complexity, substantial computational overhead, and limited scalability to large-scale scientific problems. As the number of possible research pathways grows, current methods struggle to efficiently balance exploration and exploitation, often leading to diminishing returns in both hypothesis quality and resource utilization. \textbf{(2) Autonomous learning.} A crucial component of evolution is the ability to learn from feedback, whether derived from internal self-reflection or from external sources. However, when relying solely on self-generated feedback through internal reflection, AI Scientists are prone to ``looping errors'', which means that mistakes can be amplified over multiple iterations, rather than corrected or improved. Without effective mechanisms for robust self-criticism, the system may fall into cycles of compounding errors, undermining both the originality and reliability of its research output. While incorporating feedback from external sources (especially from other AI Scientist systems) holds great promise for collective intelligence, current frameworks lack standardized communication protocols for scientist-to-scientist interactions. The absence of robust collaboration mechanisms results in inefficient exchange of ideas and suboptimal integration of external criticism, making AI Scientist systems unable to fully leverage the potential of agent collaboration, thus slowing down their evolution~\citep{wang2025rlver}.

\input{sub_tex/ethical}

%% file: sub_tex/ethical.tex
\subsection{Ethical Challenges}
\label{subsec:ethical}

The emergence of AI Scientist systems has fundamentally reshaped the landscape of human research~\citep{king2009automation}. However, as autonomous research agents, AI Scientist systems are incapable of making ethical judgments about the societal impact of their work, and they do not self-regulate based on potential risks associated with their findings~\citep{bengio2025superintelligent}. In the absence of proper oversight, AI Scientist systems may present a range of ethical challenges that require systematic consideration: 

\textbf{(1) An AI Scientist may be misused, overwhelming the peer review system and leading to a decline in overall research quality.} The rapid and large-scale generation of scientific artifacts by AI Scientist systems can flood existing peer review mechanisms, making it difficult for human reviewers to rigorously assess the quality of submissions. This overload risks allowing low-quality, erroneous, or redundant work to enter the scientific record, thereby undermining the standards of academic publishing and slowing scientific progress. Furthermore, the peer review process itself may be manipulated through techniques such as reward hacking, further diminishing the fairness and objectivity of scientific evaluation~\citep{lu2024ai}.

\textbf{(2) An AI scientist may autonomously enter dangerous research domains, accelerating the development of harmful technologies.} Without robust ethical constraints, AI Scientist systems can independently generate and publish research in sensitive areas (e.g., cybersecurity). Such uncontrolled dissemination of knowledge may accelerate the development and spread of potentially harmful scientific technologies, before adequate ethical guidelines, safety protocols, or regulatory measures can be implemented~\citep{huang2022overview}. 

\textbf{(3) An AI scientist may weaken the overall quality of scientific training and education, leading to a decline in research standards and scientific literacy across all levels of human researchers. } The widespread use of AI Scientist systems throughout the research and education pipeline may encourage over-reliance on automated assistance for key tasks such as idea generation, experimental design, data analysis, and hypothesis testing. Over time, this dependence could erode critical thinking, creativity, and hands-on research skills, ultimately diminishing scientific literacy and widening gaps between institutions with differing levels of access to advanced AI tools~\citep{yamada2025ai}.

\textbf{(4) An AI scientist may introduce security vulnerabilities and research biases in the scientific research process.} For instance, AI Scientist systems may utilize verification mechanisms unfaithful to their intended design, resulting in misleading results that are difficult to detect and require substantial human oversight~\citep{jansen2025codescientist}. These issues not only waste valuable research resources but also diminish the reliability of scientific findings. In addition, an AI Scientist tends to favor research topics with abundant datasets or high potential for automation, which can incentivize funding organizations to prioritize such areas. This skews the equitable distribution of research funding and narrows the overall research landscape~\citep{yamada2025ai}. Well-funded institutions are more capable of leveraging AI Scientist systems, further exacerbating existing inequalities and increasing entry barriers for early-career researchers.

To address these risks, there is an urgent need for a \textbf{comprehensive system of generation management, ethical oversight, and quality evaluation} for AI Scientist systems~\citep{jobin2019global}. This management system should include, but not be limited to the following components:

\textbf{(1) A centralized platform to mitigate disruption to human review systems.} This platform archives AI-generated scientific outputs and develops automated detection tools (e.g., DeepReview~\citep{zhu2025deepreview}) to identify and filter low-quality content. All AI-generated outputs must be clearly labeled with information on their origin, generation methods, and involved scientific tools. Additionally, since authorship implies legal and ethical responsibilities that current AI systems cannot assume, AI Scientist systems must not be listed as official authors, and proper attribution of legal and ethical responsibilities must be ensured.

\textbf{(2) Clear boundaries between human-driven and AI-driven research activities to preserve essential human training and ensure well-rounded development for all researchers.} 
Key stages of scientific education and training (e.g., hypothesis formulation and experimental validation) should emphasize active human involvement to uphold high scientific standards. Educational institutions and research organizations should develop clear guidelines to prevent excessive dependence on AI Scientist systems at all levels, from undergraduate students to experienced researchers, ensuring that AI serves as a tool rather than a human replacement in the educational experience.

\textbf{(3) A global convention for ethical boundaries and risk management in AI-driven research, formulating global ethics and responsibility conventions.} Full disclosure of generation processes, algorithmic sources, and potential risks should be mandatory~\citep{huang2022overview}. In addition, a hybrid oversight framework that integrates both automated systems and human-in-the-loop review should be established to provide ongoing ethical supervision and risk assessment, thereby ensuring that the research activities of AI Scientist systems remain within socially acceptable boundaries~\citep{jobin2019global,khan2022ethics}.

%% file: sub_tex/future_and_conclusion.tex
In the previous sections, we reviewed the current achievements of AI Scientist systems~(Section~\ref{sec:knowledge_acquisition}~-~Section~\ref{sec:evolution}) and analyzed their limitations~(Section~\ref{sec:discussion}) from two perspectives~(deficiencies in foundation models and inadequacies in research capabilities), highlighting where we currently stand and what remains lacking. In this section, we first outline potential directions to bridge the existing gaps in AI Scientist systems and discuss feasible pathways for their future development.

\subsection{Potential Directions to Bridge the Current Gap}
By systematically addressing foundational limitations, managing long-term research endeavors comprehensively, and developing robust communication protocols, future AI Scientist systems can progressively bridge existing gaps and realize their full potential as transformative agents in scientific discovery.

\textbf{Addressing fundamental model limitations and enhancing research abilities. }
As outlined in the previous discussion, a significant bottleneck for current AI Scientist systems lies in the intrinsic limitations of the foundation models. Issues such as hallucinations, inefficiencies in knowledge updating, and catastrophic forgetting severely impede their reliability and scientific rigor. Addressing these fundamental constraints requires \textbf{substantial advancements in model architecture, training methodologies, and inference-time verification strategies.} Potential solutions include developing hybrid model architectures that integrate symbolic reasoning capabilities with deep learning architectures to enhance interpretability and factual accuracy. Additionally, methods such as RAG and continual learning frameworks could mitigate issues related to data recency and model adaptability, ensuring that AI Scientists maintain up-to-date and verifiable scientific knowledge. Moreover, it is also crucial to \textbf{advance AI Scientist systems' research capabilities}~(e.g., hypothesis assessment, and scientific reasoning) through evolutionary mechanisms. Enhancements could include incorporating iterative preference training and structured self-reflection mechanisms, allowing models to autonomously refine their capabilities based on rigorous internal critiques and external feedback.

\textbf{Managing long-term research cycles. }
A mature AI Scientist should autonomously manage comprehensive, long-term research cycles. Current AI Scientist systems, however, predominantly focus on short-term individual research tasks. Future AI Scientist systems must incorporate sophisticated dynamic planning methods, such as \textbf{hierarchical planning frameworks} or \textbf{agentic tree-search algorithms}, to effectively navigate extensive research pathways. Additionally, to further facilitate iterative improvement loops, AI Scientist systems could incorporate frameworks like \textbf{genetic algorithm-based refinement} or \textbf{reinforcement learning-driven optimization.}
These mechanisms can help prioritize research objectives, efficiently allocate resources, and adaptively refine strategies based on progressive insights, thus significantly improving the depth and breadth of autonomous scientific discoveries.

\textbf{Developing communication protocols for AI Scientist systems. }
Another critical future direction involves developing standardized Scientist-to-Scientist Communication Protocols (SSCP), enabling structured and efficient interactions among AI Scientist systems. Current efforts in AI-driven research collaborations primarily rely on ad hoc communication methods, thus lacking standardized interaction workflows and unified communication formats. Establishing an SSCP could profoundly enhance AI Scientists' collaborative capabilities, promoting effective information exchange, hypothesis validation, and multi-agent research coordination. Such a communication protocol should specify interaction workflows, structured data formats, and collaborative research strategies to facilitate seamless integration among heterogeneous AI Scientist systems. Leveraging these standardized communication protocols could further advance the formation of AI-driven research communities, enabling collective intelligence to tackle complex interdisciplinary research challenges with unprecedented effectiveness and rigor.

\subsection{The Pathway of AI Scientist}

Unlike human scientists, \textbf{AI Scientist systems evolve through a self-organizing progression} that could be more distinct and efficient than human approaches. While human scientists accumulate knowledge through periods of formal education, hands-on experiments, and ethical training, AI Scientist systems progress by integrating specialized modules, such as LLMs, simulation engines, and robotic systems. This modular approach positions the development of AI Scientists closer to system engineering, facilitating rapid enhancements in research capabilities and enabling significant efficiency gains in scientific endeavors. Furthermore, \textbf{forcing AI Scientists to replicate human developmental paths might constrain their potential.} AI systems excel in rapidly processing and analyzing extensive datasets, enabling them to identify research problems and formulate solutions within significantly shorter timescales. Hence, development efforts for AI Scientist systems should prioritize leveraging their intrinsic strengths~(e.g., cross-disciplinary knowledge integration), rather than replicating the pathways of human scientists.

At this early stage of development, the future pathways of an AI Scientist can be envisioned along \textbf{two interdependent paradigms: (1) Personalized AI Scientist systems} tailored to individual human researchers, empowering them through personalized, co-evolutionary partnerships, and \textbf{(2) AI Scientist systems serving broader human society}, accelerating global solutions. The former aims to enhance the research productivity and creative capacities of individual scientists through tailored collaborations. In contrast, the latter seeks to establish AI-driven scientific ecosystems, effectively addressing complex global challenges. Together, these pathways converge toward a vision where AI Scientist systems transcend tool-like functionality to become proactive stewards of scientific advancement, balancing autonomy with human oversight to uphold integrity, inclusivity, and societal benefit~\citep{tsvetkova2024new}.

\subsubsection{AI Scientist for Individual Researcher}

\textbf{Active Learning.}
A personalized AI Scientist should actively gain experiences, analyze mistakes, and absorb the latest knowledge tailored for individual human researchers. This includes:  (1) Knowledge gap identification: Identifying gaps in a researcher's knowledge by reviewing their publications, notes, and unresolved questions. (2) Personalized literature curation: Recommending papers based on the researcher’s specific research interests, current project stage, and experimental bottlenecks~\citep{he2024pasa}. (3) Real-time learning: Real-time summarization of emerging literature~\citep{agarwal2024litllm} coupled with contextualized explanations of unfamiliar concepts.

\textbf{Preference alignment of human researchers. } Effective alignment involves matching the capabilities of AI Scientist systems with human research goals, including but not limited to: (1) Value-driven hypothesis generation: aligning hypothesis proposals with the researcher’s research preferences, risk tolerance and resource constraints. (2) Ethical boundary management: Dynamically adapt to the researcher’s institutional rules, funding ethics, and personal boundaries (e.g., auto-rejecting high-risk topics like gene editing). (3) Collaborative research agendas: Planning research directions with human researchers jointly, taking into account individual career aspirations and priorities for impactful scientific contributions.

\textbf{Mutual guidance and co-evolution. }
A personalized AI Scientist progresses together with human researchers through continuous scientific research activity: (1) Bidirectional refinement loops: AI Scientist systems and researchers iteratively refine scientific artifacts through mutual feedback~\citep{romera2024mathematical}. (2) Evolutionary personalization: Continuous preference tuning via reinforcement learning from human feedback~\citep{ouyang2022training}, where AI Scientist systems adapt to a researcher’s evolving style, balancing autonomy (e.g., automated verification) with human oversight.

\subsubsection{AI Scientist for Human Society}

\textbf{Adapt to the trend of scientific development.}
An AI Scientist serving broader human society must dynamically adapt to the accelerating pace and interdisciplinary nature of modern research by: (1) Continuous knowledge integration: Efficiently integrating emerging research from diverse fields (e.g., bioinformatics, quantum computing) while preventing loss of previous knowledge~\citep{kirkpatrick2017overcoming}. (2) Cross-domain collaboration: Enable interdisciplinary discovery by transferring methods across fields (e.g., applying ML-driven drug discovery to materials science). (3) Predictive research: Identifying emerging scientific trends using network analysis of preprint repositories (e.g., arXiv, bioRxiv) and funding trends, guiding research focus toward critical topics like climate resilience or pandemic preparedness.

\textbf{Prospects for human development. }
AI Scientist systems can greatly boost human progress by: (1) Reducing resource gaps by providing open-access tools and cloud-based experimental simulators, enabling under-resourced institutions to participate in global research. (2) Accelerating discovery for critical challenges (e.g., renewable energy, disease treatment) through automated experimentation. Early successes like ChemCrow~\citep{bran2023chemcrow} and BioDiscoveryAgent~\citep{roohani2024biodiscoveryagent} show promise in chemistry/biology.
As these systems increasingly interface with physical laboratories (e.g., IoT sensors), they may evolve to support full-cycle scientific workflows.